\documentclass[10pt,twocolumn,twoside]{IEEEtran}

\usepackage{graphicx}
\usepackage{amsmath,epsfig,amssymb,amsbsy,bm}

\usepackage{algpseudocode}

\usepackage{algorithm}

\usepackage{epstopdf}
\usepackage{multirow}
\graphicspath{{images/}}
\usepackage[x11names]{xcolor}

\usepackage[normalem]{ulem}
\newtheorem{theorem}{Theorem}[section]

\newtheorem{proposition}[theorem]{Proposition}

\usepackage{soul}

\newcommand{\qed}{\nobreak \ifvmode \relax \else
      \ifdim\lastskip<1.5em \hskip-\lastskip
      \hskip1.5em plus0em minus0.5em \fi \nobreak
      \vrule height0.75em width0.5em depth0.25em\fi}

\begin{document}
\title{Attentive monitoring of multiple video streams driven by a  Bayesian foraging strategy}
\author{Paolo~Napoletano,~\IEEEmembership{Member,~IEEE,}  Giuseppe~Boccignone  and Francesco~Tisato
\thanks{P. Napoletano and F. Tisato are with the Dipartimento di Informatica, Sistemistica e Comunicazione, Universit\'a degli Studi Milano--Bicocca , Viale Sarca 336, 20126 Milano, Italy.}
\thanks{G. Boccignone is with the  Dipartimento di Informatica, Universit\'a degli Studi Milano Statale, Via Comelico 39/41
Milano, 20135 Italy.}
}

\markboth{IEEE xxxx}{Napoletano \MakeLowercase{\textit{et al.}}:Multi-camera}

\maketitle
\begin{abstract}

In this paper we shall consider the  problem of deploying attention to  subsets of the video streams for  collating the most relevant data and information of interest related to a given task. 
We formalize this monitoring problem as a foraging problem. We propose a  probabilistic framework to  model observer's attentive behavior as the behavior of a forager. The forager, moment to moment, focuses  its attention on the most informative stream/camera, detects interesting objects or activities, or switches  to a more profitable stream.

The approach proposed here is suitable to be exploited  for multi-stream video summarisation. Meanwhile, it  can serve as a preliminary step for more sophisticated video surveillance, e.g. activity and behavior analysis. 
Experimental results achieved on the UCR Videoweb Activities Dataset, a publicly available dataset, are presented to illustrate the utility of the proposed technique.
 \end{abstract}


\IEEEpeerreviewmaketitle

\section{Introduction}
\label{sec:intro}

\IEEEPARstart{T}{he} volume of data collected by current networks of cameras for video surveillance  clearly overburdens the monitoring ability of human viewers to stay focused on a task. Further, much of the data that can be collected from multiple video streams  is uneventful.  Thus, the need for the discovery and the selection of activities occurring within and across videos for  collating  information  most relevant to  the given task has fostered the field of  multi-stream summarise.   

At  the heart of multi-stream summarisation there is a ``choose  and leave'' problem that moment to moment an ideal or optimal observer (say, a software agent)  must solve:  choose the   most informative stream; detect, if any, interesting  activities occurring within the current stream; leave  the  handled stream for  the next ``best'' stream.

In this paper, we provide a different perspective to such ``choose  and leave'' problem based on a principled framework that unifies  overt visual attention behavior  and optimal foraging. The framework  we propose is
just one, but a novel, way of formulating the multi-stream summarisation problem and solution (see Section \ref{sec:relwork}, for a discussion). 

In a nutshell, we consider  the  foraging landscape of multiple streams, each video stream being  a \emph{foraging patch}, and the ideal observer playing the role of the visual forager (cfr. Table \ref{metaphor}). According to Optimal Foraging Theory (OFT), a forager that feeds on patchily distributed preys or resources, spends its  time traveling between patches or searching and handling food within patches \cite{stephens1986foraging}. While  searching, it gradually depletes the food, hence, the benefit of staying in the patch is likely to gradually diminish with time.  Moment to moment, striving to maximize its foraging efficiency and energy intake, the forager should make decisions: Which is the best patch  to search?    Which prey, if any, should be chased within the patch? When to leave the current  patch for a richer one?

Here   visual foraging corresponds to the  time-varying overt deployment of  visual attention achieved through oculomotor actions, namely, gaze shifts. Tantamount to the forager, the observer is pressed to maximize his information intake over time under a given task, by moment-to-moment sampling the most informative subsets of video streams. 
\begin{table}[htdp]
\scriptsize
\caption{Relationship between Attentive vision and Foraging}
\begin{center}
\begin{tabular}{|c|c|}
\hline
\textbf{Multi-stream attentive processing} & \textbf{Patchy landscape foraging} \\
\hline
Observer & Forager \\
\hline
Observer's gaze shift & Forager's relocation \\
\hline
Video stream  & Patch\\
\hline
Proto-object & Candidate prey\\
\hline
Detected object & Prey\\
\hline
Stream  selection & Patch choice \\
\hline
Deploying attention to object  & Prey choice and handling\\
\hline
Disengaging from object  & Prey leave\\
\hline
Stream leave & Patch leave or giving-up \\
\hline
\end{tabular}
\end{center}
\label{metaphor}
\end{table}%
All together, choosing  the ``best'' stream,  deploying attention to within-stream activities,  leaving the  attended stream, represent the unfolding of a dynamic decision making process. Such monitoring decisions have to be made by relying upon automatic interpretation of scenes for  detecting   actions and activities.  To be consistent with the terminology proposed in the literature \cite{xiang2006beyond},  an \emph{action} refers to a sequence of movements executed by a single \emph{object} ( e.g.,  ``human walking'' or ``vehicle turning right'').  An \emph{activity}  contains a number of sequential actions, most likely involving multiple objects  that interact or co-exist in a shared common space monitored by single or multiple cameras (e.g., ``passengers walking on a train platform and sitting down on a bench''). The ultimate goal of activity modelling is to understand \emph{behavior}, i.e. the meaning of activity in the shape of a semantic description. Clearly,   action/activity/behavior analysis entails the capability of spotting objects that are of interest for the given surveillance task.

Thus, in the work presented here the visual objects of interest occurring in  video streams are the \emph{preys} to be chased and handled  by the visual forager. 
 Decisions at the finer level of  a single stream concern which object is to be chosen and analyzed (\emph{prey choice and handling},  depending on task), and when to disengage from the spotted object for deploying attention to the next (\emph{prey leave}).

\begin{figure}[tb]%
\centering
\includegraphics[scale=0.25]{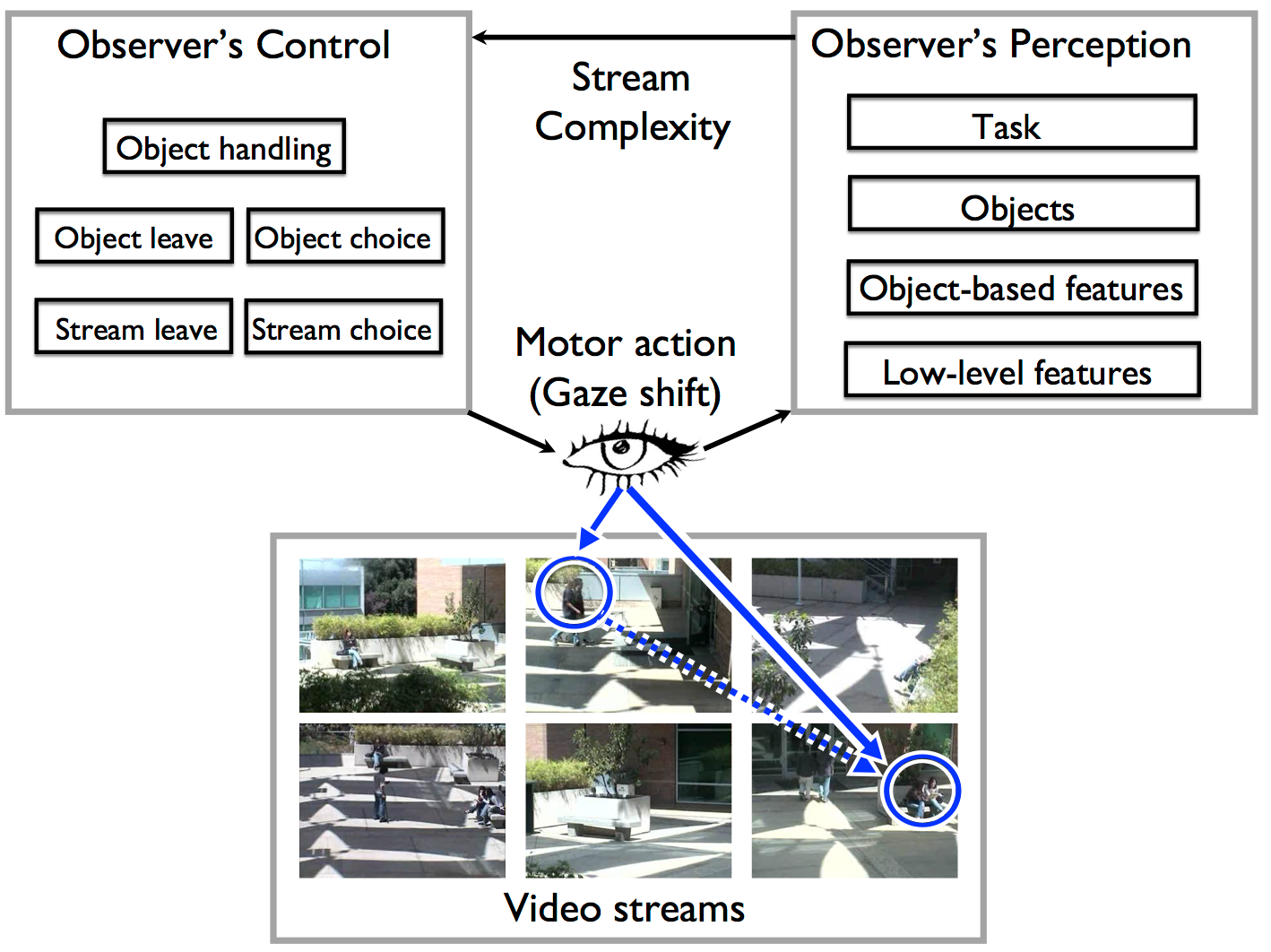}
\caption{Monitoring of multiple video streams  as attentive foraging. The ideal observer (forager) is involved in a perception/action loop that supports foraging activity.  Multiple streams are the raw sensory input. The observer pre-attentively selects the most informative stream (patch) and sets his Focus of Attention via  a gaze shift action;  within the stream, interesting objects  (preys) are attentively detected and handled through local gaze shifts. Moment to moment,  a Bayesian optimal strategy is exploited to make a decision whether to stay or  leave the scrutinized stream by shifting the gaze to a more profitable one. The strategy relies upon the perceptual feedback of the overall ``quality'' (complexity) of  streams.}%
\label{fig:model}%
\end{figure}

The reformulation of visual attention in terms of foraging theory is not simply an  informing metaphor.  What was once foraging for tangible resources in a physical space  became, over evolutionary time, foraging in cognitive space for information related to those resources \cite{hills2006animal}, and such adaptations play a fundamental role in goal-directed deployment of visual attention \cite{wolfe2013time}. 
Under these rationales, we present a  model of  Bayesian observer's  attentive foraging supported by the perception/action cycle  presented in  Fig. \ref{fig:model}.  Building on  the perception/action cycle, visual attention provides an efficient allocation and management of resources.  

The cycle embodies two main functional blocks: the perceptual component and the executive control component.
The perceptual component is in charge of ``What'' to look for, and the executive component accounts for the overt attention shifts, by deciding   ``Where and How'' to look at, i.e., the actual gaze position, and thus the observer's Focus of Attention (FoA).   The observer's perceptual system operates on  information represented at different levels of abstraction (from raw data to task dependent information); at any time, the currently  sensed visual stimuli depend on the  oculomotor action or gaze shift performed either within the stream (\emph{within-patch}) or across streams (\emph{between-patch}). Based on perceptual inferences at the different levels, the main feedback information passed on to the executive component, or controller, is an index of stream quality formalized in terms of the configurational  complexity of potential objects sensed within the stream.  

A stream is selected by relying upon its pre-attentively sensed quality. Once  within stream, the observer attentively detects and handles objects  that are informative for the given task.  Meanwhile, by intra-stream foraging, the observer  gains information on the actual stream quality  in terms of experienced detection rewards.
 Object handling within the  attended stream occurs until a decision is made to leave  for a more profitable stream. Such decision  relies upon a Bayesian strategy, which is the core of this paper. The strategy  extends to  stochastic landscapes observed under incomplete information,  the deterministic global policy derived from classic Charnov's Marginal Value Theorem (MVT,~\cite{charnov1976optimal}), while integrating within-stream observer's experience.

By relying on such perception/action cycle, we assume that the deployment of  gaze to one video frame  precisely reflects the importance of that frame.  Namely,  given  a number of video streams as the input, at any point in time, we designate the  current gazed  frame  as the relevant video frame to be included in the  final output summarisation. The output succinctly captures the most important data (objects engaged in actions) for the surveillance analysis task.

The idea  of a layered framework for the control of gaze deployment, implementing a general perception/action loop is an important one in the visual attention  literature (cfr., Sch{\"u}tz  \emph{et al.} \cite{schutz2011eye} for a discussion) and it has been fostered by Fuster~\cite{fuster2004upper,fuster2009cortex}. Such idea together with the assumption that attention is algorithmic in nature and needs not  to occupy a distinct physical place in the brain is germane to our theme. Fuster's paradigm has been more recently formalized  by Haykin~\cite{haykin2014cognitive} under the name of \emph{Cognitive Dynamic Systems}. In this perspective, our cognitive foraging approach to monitoring  can be considered closely related  to the Cognitive Dynamic Surveillance System (CDSS) approach,  a remarkable and  emerging  domain proposed by Regazzoni and colleagues \cite{chiappino2014bio,chiappino2014event,dore2010interaction}. In CDSS,  attentive mechanisms \cite{Regazzoni_AVSS2013} are likely to add relevant value in the effort of  designing the next generation of surveillance systems.  

In the rest of this paper, Section~\ref{sec:relwork} describes the related literature and contributions of this work. Section~\ref{sec:model}  provides a formal overview of the model.  
Section~\ref{sec:landsampling} details the pre-attentive stage. Stream selection is discussed in  Section~\ref{sec:patch_choice}.  Section~\ref{sec:patchsampling} describes within-stream visual attention deployment, while Section~\ref{sec:patchleave} discusses the Bayesian strategy for leaving the stream.
Experimental work is presented in Section~\ref{sec:exp}. Finally, Section~\ref{sec:final} concludes this paper.

\section{Related work and our contributions}
\label{sec:relwork}

Main efforts in the summarisation literature  have been spent on the single-camera case, while  the multi-camera setting has not received as much attention (see \cite{Wang2013,leo2014multicamera,online_2015} for review). Specifically, the work by Leo and Manjunath\cite{leo2014multicamera} shares our concern of providing a unified framework to generate summaries. Different from us, they rely on document analysis-inspired activity motif discovery. Time series of  activations are computed from dense optical flow in different regions of the video and high-level activities are identified  using the  topic model analysis. The step from activities detected in individual video streams to a complete network summary relies on identifying  and reducing inter-and intra-activity redundancy. This approach, cognate with those based on sparse coding dictionaries for finding the most representative frames (e.g.,\cite{cong2012towards}), requires off-line learning of activities from documents, each document being the time series of activations of a single stream. While offering some advantage for inferring high level activities\cite{Wang2013},  these methods avoid confronting with complex  vision problems and distributed optimal control strategies  brought on by the multi-stream setting \cite{tron2011distributed,MichelForesti2010,Roy-Chow_review2014}. On the other hand,  difficulties arise when dealing with large video corpora and with dynamic video streams,  e.g. on-line summarisation in visual sensor networks \cite{online_2015}, a case which is more related to our scenario.

In this view, beyond multi-camera summarisation,   it is of interest work concerning multi-camera surveillance, where manual  coordination  becomes unmanageable when the number of cameras is large.  To some extent, the ``choice and leave'' problem previously introduced  bears relationships with 
two challenging issues:  camera assignment  (which camera is being used to extract essential information)
 and camera handoff (the process of finding the next best camera). Indeed,   the  complexity of these problems on large networks is such that    Qureshi and Terzopoulos~\cite{qureshi2008smart} have proposed  the use of virtual environments  to demonstrate camera selection and handover strategies.  Two remarkable papers address the issue of designing a general  framework inspired by non-conventional theoretical analyses, in a vein similar to the work presented here. Li and Bhanu~\cite{li2011utility} have presented an approach based on game-theory.  Camera selection is based on a utility function that is computed by a bargaining  among cameras capturing the tracking object.
 Esterle \emph{et al.}~\cite{esterle2014socio} adopted  a fully decentralised socio-economic approach for online  handover in smart camera networks. Autonomous cameras exchange responsibility for tracking objects in a market mechanism in order to maximize their own utility. When a handover is required, an auction is initiated and cameras that have received the auction initiation try to detect the object within their the field of view. 

At this point it is worth noting that, in the effort towards a general framework for stream selection and handling, all works above, differently from the approach we present here,  are quite agnostic about the image  analysis techniques to adopt.  They mostly rely on basic  tools (e.g., dense optical flow \cite{leo2014multicamera}, Camshift tracking manually initialized~\cite{li2011utility}, simple frame-to-frame SIFT computation~\cite{esterle2014socio}). However, from a general standpoint, moving object detection and recognition, tracking, behavioral analysis  are  stages that deeply involve the realms of image processing  and  machine vision.  
In these research areas, one major concern that has been an omnipresent topic during the last years  is how to restrict the large amount of visual data to a manageable rate \cite{MichelForesti2010,tron2011distributed}. 

Yet to tackle information overload, biological vision systems have evolved a remarkable capability:  visual attention,  which gates relevant information to subsequent complex processes (e.g., object recognition).
A series of studies published under the headings of Animate~\cite{Ballard}, or  Active  Vision~\cite{aloimonos2013active} has investigated how the concepts of human selective attention can be exploited for computational systems dealing with a large amount of  image data (for an extensive review, see  \cite{BorItti2012}). Indeed,   determining the most interesting regions of an image in a ``natural'', human-like way is a promising approach to improve computational vision systems. 

Surprisingly enough, the issue of attention has been hitherto overlooked by most approaches  in  video surveillance, monitoring and summarisation~\cite{Wang2013,online_2015}, apart from those in the emerging domain  of smart camera networks embedding  pan-tilt-zoom (PTZ) cameras.  PTZ cameras can actively change intrinsic and extrinsic parameters to adapt their field of view (FOV) to specific tasks~\cite{MichelForesti2010,Roy-Chow_review2014}. In such domain, active vision is a pillar~\cite{MichelForesti2010,tron2011distributed}, since FOV adaptation can be
exploited to focus the ``video-network attention'' on areas of interest. In PTZ  networks, each of the cameras is assumed to have its own embedded target detection module, a distributed tracker that provides an estimate of the state of each target in the scene, and  a distributed camera control mechanism~\cite{Roy-Chow_review2014}.  Control issues have been central to this field: the large amount of camera nodes in these networks and the tight resource limitations requires balancing among conflicting goals \cite{micheloni_TCSVT2011,qureshi2008smart}. In this respect, the exploitation vs. exploration dilemma is cogent here much like in our work. For example, Sommerlade and Reid \cite{sommerlade2010probabilistic} present a probabilistic approach to  maximize the expected mutual information gain as a measure for the utility of each parameter setting and task. The approach allows balancing conflicting objectives such as target detection and obtaining high resolution images of each target. Active distributed optimal control  has been given a Bayesian formulation in a game theoretic setting. The Bayesian formulation enables automatic trading-off of objective maximization versus the risk of losing track of any target; the game-theoretic design allows the global problem to be decoupled into local problems at each PTZ camera~\cite{Roy-Chow_TIP2012,Roy-Chow_CST2014}.

In most cases visual routines and control are treated as related but technically distinct problems~\cite{Roy-Chow_review2014}. Clearly, these involve a number of fundamental challenges to the existing technology in computer vision  and the quest for efficient and scalable distributed  vision algorithms \cite{tron2011distributed}.
The primary goal of these systems has been   tracking  distinct targets, where adopted  schemes are  extensions of the classic Kalman Filter to the distributed estimation framework~\cite{Roy-Chow_review2014}.  However, it is important to note that tracking is but one aspect of multi-stream analysis and of visual attentive behavior (\cite{xiang2006beyond}, but see Section \ref{sec:action_contr} for a discussion).  To sum up, while the development of PTZ networks has cast interest for  active vision techniques that are at the heart of the attentive vision paradigm~\cite{Ballard,aloimonos2013active}, yet even in this field we are far from a full exploitation of tools made available by such paradigm.   

There are some exceptions to this general state of affairs. The use of visual attention   has been proposed by Kankanhalli \emph{et al.}\cite{kankanhalli2006expB}. They embrace the broad perspective of multimedia data streams, but the stream selection process is yet handled within the classic framework of optimization theory and relying on an attention measure (saturation, \cite{kankanhalli2006expB}).  Interestingly, they resort  to the MVT result, but only for experimental evaluation purposes. In our work  the Bayesian extension of the MVT is at the core of the process.    The interesting work by    Chiappino \emph{et al.} ~\cite{Regazzoni_AVSS2013} proposes a bio-inspired algorithm for attention focusing on densely populated areas and for detecting anomalies in crowd. Their  technique relies on an entropy measure and in some respect bears some resemblance to  the pre-attentive  monitoring stage of our model. Martinel \emph{et al.} \cite{martinel_micheloni_2014saliency} identify the salient regions of a given person, for person re-identification across non-overlapping camera views. 
 Recent work on video summarisation has borrowed salience representations from the visual attention realm. Ejaz \emph{et al.} \cite{ejaz2013efficient} choose key frames  as  salient frames on the basis of  low-level salience.  High-level salience based on most important objects and people is exploited in \cite{lee2012discovering} for summarisation, so that the storyboard frames  reflect the key object-driven events. Albeit not explicitly dealing with salience, since building upon sparse coding summarisation , Zhao and Xing \cite{zhao2014quasi} differentiate from \cite{cong2012towards} and generate  video summaries   by combining segments that \emph{cannot} be reconstructed using the learned dictionary. Indeed, this approach, which  incorporates in summaries unseen and interesting contents, is equivalent to denote salient those events that are unpredictable on  prior knowledge (salient as ``surprising'', \cite{BorItti2012}). Either \cite{lee2012discovering} and  \cite{zhao2014quasi} only consider single-stream summarisation.
The use of high-level saliency to handle the multi-stream  case has been addressed in \cite{napoletano2014attentive}, hinging on \cite{bocc08tcsvt}; this method can be considered as a baseline deterministic solution  to the  problem addressed here (cfr., for further analysis, Section~\ref{sec:exp}).

   Our method is fundamentally different from all of the above approaches. We work within the  attentive framework but the main novelty is that by focusing on the gaze as the principal paradigm for active perception, we reformulate the deployment of gaze to a video stream or to objects within the stream as a stochastic foraging problem. This way we unify intra- and inter-stream analyses.
 More precisely, the main technical contributions of this paper lie in the following. 

First,  based on OFT, a stochastic extension of  the MVT  is proposed,  which defines an optimal strategy for a Bayesian visual forager. The strategy combines in a principled way global information from the landscape of streams with local information gained in attentive within-stream analysis.    
The  complexity measure that is used   is apt to be exploited for within-patch analysis (e.g, from group of people to single person behavior), much like some foragers do by exploiting a hierarchy of patch aggregation levels \cite{waage1979foraging}. 

Second,  the visual attention problem is formulated as a foraging problem by extending  previous work  on  L\'evy flights as a prior for sampling gaze shift amplitudes   \cite{BocFerSMCB2013}, which mainly relied on bottom-up salience.  At the same time, task dependence is introduced, which is not achieved through ad hoc procedures. It  is naturally integrated within attentional mechanisms in terms of rewards experienced in the attentive stage when the stream is explored. This  issue  is seldom taken into account in computational models of visual attention (see \cite{BorItti2012,schutz2011eye} but in particular Tatler \emph{et al} \cite{TatlerBallard2011eye}). A preliminary study on this challenging problem has been presented in \cite{BocCOGN2014}, but  limited to the  task of searching for text in static images. 

\section{Model overview}
\label{sec:model}
In this Section we present an overview of the model  to frame  detailed discussion of its key aspects  covered in Sections \ref{sec:landsampling} (pre-attentive analysis),  \ref{sec:patch_choice} (stream choice), \ref{sec:patchsampling} (within-stream attentive analysis) and \ref{sec:patchleave} (Bayesian strategy for stream leave).

Recall from Section \ref{sec:intro} that the input to our system is a visual \emph{landscape} of $K$   video streams, each stream being a sequence of time parametrized  frames $\{\mathbf{I}^{(k)}(1),\mathbf{I}^{(k)}(2), \cdots,\mathbf{I}^{(k)}(t),\cdots \}$, where  $t$ is the time parameter and $k \in \left[1,\cdots, K\right]$. Denote $\mathcal{D}$ the spatial support of $\mathbf{I}^{(k)}$, and $\mathbf{r}^{(k)} \in \mathcal{D}$   the coordinates of a point in such domain. By relying  on the perception/action cycle outlined in Fig. \ref{fig:model}, at any point $t$ in time, we designate the  current gazed  frame $\mathbf{I}^{(k)}(t)$ of  stream $k$ as the relevant video frame to be  selected and included in the  final output summarisation

To such end,   each video stream is  the equivalent of a foraging patch (cfr. Table \ref{metaphor})  and   objects of interest (preys)   occur within the stream.
In OFT terms, it is assumed that: the landscape is stochastic; the forager has sensing capabilities and it can gain information on patch quality and available preys as it forages. Thus, the model is conceived in a probabilistic framework. Use the following random variables (RVs):
\begin{itemize}
\item $\mathbf{T}$: a  RV with $|\mathbf{T}|$ values corresponding to the \emph{task} pursued by the observer.
\item $\mathbf{O}$: a multinomial RV with $|\mathbf{O}|$ values corresponding to \emph{objects}  known by the observer
\end{itemize} 

As a case study, we  deal with actions and activities involving people.  Thus, the given task $\mathbf{T}$ corresponds to ``pay attention to people within the scene''.  To this purpose, the classes of objects of interest for the observer are represented by faces and human bodies, i.e., $\mathbf{O}=\{face, body\}$. 

The  observer  engages in a  perception/action  cycle to accomplish the given task (Fig.\ref{fig:model}). Actions are represented by the moment-to-moment relocations of gaze, say $ \mathbf{r}_{F}(t-1) \mapsto \mathbf{r}_{F}(t)$, where $ \mathbf{r}_{F}(t-1)$ and $ \mathbf{r}_{F}(t)$ are  the old and new  gaze positions, respectively.  We deal with two kinds of relocations: i) from current video stream $k$ to the next selected $k'$ (between-patch shift), i.e. $ \mathbf{r}^{(k)}_{F}(t-1) \mapsto \mathbf{r}^{(k^\prime)}_{F}(t)$; ii) from one  position to another within  the selected  stream (within-patch gaze shifts), $ \mathbf{r}^{(k)}_{F}(t-1) \mapsto \mathbf{r}^{(k)}_{F}(t)$. Since we assume unitary time for between-stream shifts, in the following we will drop the $k$ index and  simply use $\mathbf{r}_{F}$ to denote the center of the FoA  within the frame without ambiguity.   
Relocations occur because of decisions taken by the  observer upon his own perceptual inferences. In turn, moment to moment, perceptual inferences are conditioned on the observer's current FoA set by the gaze shift action. 

\subsection{Perceptual component}
\label{sec:sensing}
Perceptual inference stands on the visual features that can be extracted from raw data streams, a feature being a function $f : \mathbf{I}(t) \rightarrow F_{f}(t)$. In keeping with the visual attention literature \cite{schutz2011eye}, we distinguish between two kinds of features:
\begin{itemize}
\item \emph{bottom-up} or feed-forward features, say $\mathbf{F}_{|\mathbf{I}}$ - such as edge, texture, color, motion features - corresponding to those  that  biological visual systems learn along evolution  or in early development stages  for identifying sources of stimulus information available in the environment (\emph{phyletic} features, \cite{fuster2009cortex});

\item \emph{top-down} or object-based features, i.e. $\mathbf{F}_{|\mathbf{O}}$.
\end{itemize}

There is a large variety of bottom-up features that could be used (see \cite{BorItti2012}). Following \cite{seo2009}, we first compute, at each point $\mathbf{r}$ in the spatial support of the frame $\mathbf{I}(t)$ from the given stream, spatio-temporal first derivatives (w.r.t temporally adjacent frames $\mathbf{I}(t-1)$ and $\mathbf{I}(t+1)$). These are exploited to estimate, within a  window,  local covariance matrices $\mathbf{C}_{\mathbf{r}} \in \mathbb{R}^{3 \times 3}$, which in turn are used to compute  space-time local steering kernels 
$K(\mathbf{r} -\mathbf{r}^{'}) \propto \exp \{\frac{(\mathbf{r} -\mathbf{r}^{'})^{T}\mathbf{C}_{\mathbf{r} }  (\mathbf{r} -\mathbf{r}^{'})}{-2h^2} \}$. Each kernel response is vectorised as $\mathbf{f}_{\mathbf{r}}$. Then vectors $\mathbf{f}_{\mathbf{r}}$ are collected 
 in a local window ($3 \times 3$) and in a center + surround window ($5 \times 5$) both centered   at $\mathbf{r}$ to form a feature matrix $\ \mathbf{F}^{(k)}_{|\mathbf{I}}$. The motivation for using such features stems from the fact that local regression kernels capture the underlying local structure of the data exceedingly well, even in the presence of significant distortions. Further they do not require explicit motion estimation.
 
As to object-based features, these are to be learned by specifically taking into account the classes of objects at hand.  In the work presented here, the objects of interest are  $\mathbf{O}=\{face, body\}$; thus,  we compute face and person features by using  the Haar/AdaBoost features exploited by the well-known Viola-Jones detector. This is  a technical choice guided by computational efficiency issues; other choices \cite{Poggio2010,bocc08tcsvt} would be equivalent from the modeling standpoint. 

In order to be processed, features need to be spatially organized in feature maps. A feature map $\mathbf{X}$ is a topographically organized map that encodes the joint occurrence of a specific feature at a spatial location. It can be equivalently represented as  a unique  map encoding the presence of different object based features $\mathbf{F}^{(k)}_{f|\mathbf{O}}$ (e.g., face and body map), or a set of object-specific feature maps, i.e. $\mathbf{X}=\{\mathbf{X}_{f}\}$ (e.g., a face map, a body map, etc.).  More precisely, referring to the $k$-th stream,   $\mathbf{X}_{f}^{(k)}(t)$ is   a matrix of  binary RVs  $x_{f}^{(k)}(\mathbf{r}, t)$   denoting if feature $f$ is present or not present at location $\mathbf{r}$ at time $t$. Simply put, given $f$,  $\mathbf{X}_{f}^{(k)}(t)$ is a map defining the spatial mask of $\mathbf{F}^{(k)}_{f|\mathbf{O}}$. 

To support gaze-shift decisions, we define the RV $\mathbf{L}$ capturing the  concept of priority map. Namely, for the $k$-th stream, denote $\mathbf{L}^{(k)}(t)$ the  matrix of  binary RVs  $l^{(k)}(\mathbf{r}, t)$ denoting if  location $\mathbf{r}$ is to be considered relevant ($l^{(k)}(\mathbf{r}, t)=1$) or not ($l^{(k)}(\mathbf{r}, t)=0$) at time $t$. It is important to note that the term ``relevant'' is to be specified with respect to the kind of feature map used to infer a probability density function (pdf) over $\mathbf{L}^{(k)}$. For instance,  if only bottom-up features are taken into account, then ``relevant'' boils down to ``salient'', and gaze shifts will be driven by the physical properties of the scene, such as motion, color, etc. 

Eventually, in accordance with object-based attention approaches, we introduce \emph{proto-objects}  $\mathcal{O}^{(k)}(t)$ as the actual dynamic support for gaze orienting. Following Rensink \cite{rensink2000dynamic}, they are conceived as the dynamic interface between attentive and pre-attentive processing. Namely, a ``quick and dirty" time-varying perception of the scene,  from which a number of proto-objects is suitable to be glued in the percept of an object by the attentional process.  Here, operatively,  proto-objects are drawn from the priority map and each proto-object is used to sample \emph{interest points} (IPs). The latter  provide a sparse representation of a candidate objects to gaze at; meanwhile, the whole set of IPs sampled at time $t$ on video stream $k$ is used to compute the  configurational complexity, say  $\mathcal{C}^{(k)}(t)$, which is adopted as a prior quality index of the stream, in terms of foraging opportunities (potential preys within the patch).
 
More generally, whilst  $\mathbf{X}$ and $\mathbf{L}$   can be conceived as perceptual memories, the dynamic ensemble of proto-objects is  more similar to a  working memory that allows attention to be temporarily focused on an internal representation~\cite{fuster2009cortex}.

\begin{figure}[tb]%
\centering
\includegraphics[scale=0.20]{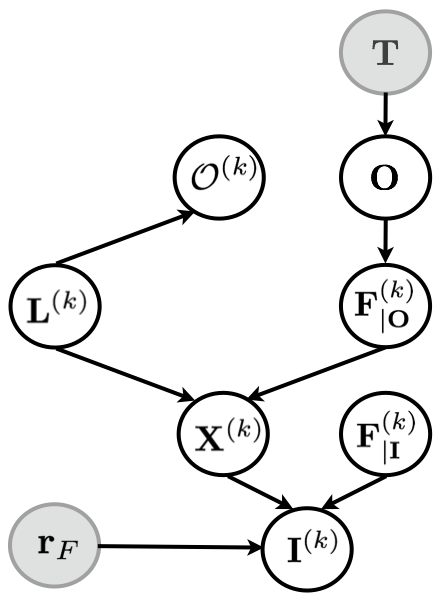}
\caption{The perception component (cfr. Fig. \ref{fig:model})   as a Probabilistic Graphical Model. Graph nodes denote RVs and  directed arcs  encode conditional dependencies between RVs. Grey-shaded  nodes  stand for RVs whose value is given (current gaze position and task). Time index $t$  has been omitted for simplicity}%
\label{fig:sense}%
\end{figure}

Given the task $\mathbf{T}$ and the current gaze position $\mathbf{r}_{F}(t)$, perceptual inference relies upon the joint pdf $P(\mathbf{O},\mathbf{F}_{|\mathbf{O}}^{(k)}(t),  \mathbf{L}^{(k)}(t), \mathcal{O}^{(k)}(t), \mathbf{X}^{(k)}(t),  \mathbf{F}_{|\mathbf{I}}, \mathbf{I}^{(k)}(t)  |  \mathbf{T}, \mathbf{r}_{F}(t))$.   The representation of such pdf can be given the form of  the directed Probabilistic Graphical Model (PGM, \cite{koller2009probabilistic}), say $\mathcal{G}$,  presented in Fig. \ref{fig:sense}. 
The PGM  structure captures the assumptions about the visual  process previously discussed.  For example,  the assumption that given  task  $\mathbf{T}$,  object class $\mathbf{O}$ is likely to occur,  is represented through the dependence $\mathbf{T} \rightarrow \mathbf{O}$. 

Stated technically, the $\mathcal{G}$ structure encodes  the set $\mathcal{I}_{\ell}(\mathcal{G})$ of conditional independence assumptions over  RVs (the local independencies, \cite{koller2009probabilistic}) involved by  the joint pdf. Then,  the  joint pdf factorizes according to $\mathcal{G}$  (cfr.,  Koller~\cite{koller2009probabilistic}, Theorem 3.1):
\begin{multline}
P(\mathbf{O},\mathbf{F}_{|\mathbf{O}}^{(k)},  \mathbf{L}^{(k)}, \mathcal{O}^{(k)}, \mathbf{X}^{(k)}, \mathbf{F}_{|\mathbf{I}}^{(k)}, \mathbf{I}^{(k)}  \mid  \mathbf{T}, \mathbf{r}_{F}) = \\
 P(\mathbf{O} \mid \mathbf{T}) P(\mathbf{F}_{|\mathbf{O}}^{(k)} \mid \mathbf{O}) P(\mathbf{L}^{(k)}) P(\mathcal{O}^{(k)} \mid \mathbf{L}^{(k)}) \cdot \\
  P(\mathbf{X}^{(k)} \mid \mathbf{L}^{(k)},\mathbf{F}_{|\mathbf{O}}^{(k)}) P( \mathbf{I}^{(k)} \mid \mathbf{F}_{|\mathbf{I}}^{(k)}, \mathbf{X}^{(k)}, \mathbf{r}_{F})
\label{eq:joint}
\end{multline}
\noindent (time  index $t$ has been omitted for notational simplicity).
\noindent The factorization specified in Eq. \ref{eq:joint} makes explicit the local distributions (the set of independence assertions $\mathcal{I}(P)$ that hold in pdf $P$, $\mathcal{I}_{\ell}(\mathcal{G}) \subseteq \mathcal{I}(P)$), and related inferences at the different levels of visual representation guiding gaze deployment.

\subsubsection{Object-based level}
$P(\mathbf{O} \mid \mathbf{T})$ is the multinomial distribution defining the prior on object classes under the given task,  whose parameters can be easily estimated via Maximum-Likelihood (basically, object occurrence counting).

$P(\mathbf{F}_{|\mathbf{O}}^{(k)}(t) \mid \mathbf{O})$  represents the object-based feature likelihood. In current simulation, we use the Viola-Jones detector for faces and persons and convert the outcome  to a probabilistic output (see  \cite{BocBoosted}, for a formal justification).

\subsubsection{Spatial-based level}
$P(\mathbf{L}^{(k)})$ denotes the prior probability of gazing at  location $\mathbf{L}^{(k)}= \mathbf{r}^{(k)}$ of the scene. For example,  specific pdfs can be learned  to account for the  gist of the scene \cite {TorrJOSA} (given a urban scene, pedestrian  are more likely to occur in the  middle horizontal region)  or  specific spatial biases, e.g. the central fixation bias  \cite{TatlerBallard2011eye}. Here, we will not account for such tendencies, thus we assume a uniform prior. The factor $P(\mathcal{O}^{(k)}(t)| \mathbf{L}^{(k)}(t))$ is the proto-object likelihood given the priority map, which will be further detailed in Section~\ref{sec:patch_choice}.

\subsubsection{Feature map level} 
$P(\mathbf{X}_{f}^{(k)}(t) \mid \mathbf{L}^{(k)}(t),\mathbf{F}_{|\mathbf{O}}^{(k)}(t))$ represents  the likelihood of  object-based feature $\mathbf{F}_{|\mathbf{O}} = \mathbf{f}_{|\mathbf{O}}$ to occur at location $\mathbf{L}^{(k)}= \mathbf{r}^{(k)}$. Following~\cite{Poggio2010}, when the feature is present at   $\mathbf{r}^{(k)}$  we set $P(\mathbf{X}_{f}^{(k)}(t) =1 \mid \mathbf{L}^{(k)}(t)=\mathbf{r}^{(k)},\mathbf{F}_{|\mathbf{O}}^{(k)}(t)=1)$ equal to a Gaussian $\mathcal{N}(\mathbf{r}^{(k)},\sigma)$ centered at $\mathbf{r}^{(k)}$ ($\sigma=1)$, to activate nearby locations; otherwise, to a small value $P(\mathbf{X}_{f}^{(k)}(t) =0 \mid \mathbf{L}^{(k)}(t)=\mathbf{r}^{(k)},\mathbf{F}_{|\mathbf{O}}^{(k)}(t)=0)= \epsilon$ ($\epsilon= 0.01$).

The factor $P( \mathbf{I}^{(k)}(t) \mid \mathbf{F}_{|\mathbf{I}}, \mathbf{X}_{f}^{(k)}(t), \mathbf{r}_{F}(t))$ is the feed-forward evidence obtained from low-level features $\mathbf{F}_{|\mathbf{I}}$ computed  from frame $\mathbf{I}^{(k)}(t)$ as sensed when gaze is set at $\mathbf{r}_{F}(t)$. In the pre-attentive stage the position of gaze is not taken into account, and the input frame is a low-resolution representation of the original. In the attentive stage, $\mathbf{r}_{F}(t)$ is used to simulate foveation -  accounting for  the contrast sensitivity fall-off  moving from the center of the retina, the fovea, to the periphery; thus, the input frame is a  \emph{foveated image}
~\cite{bocc08tcsvt}. 

The feed-forward evidence is proportional to the output of low-level filters $f : \mathbf{I}(t) \rightarrow F_{f}(t)$. A variety of approaches can be used \cite{BorItti2012} from a simple normalization of filter outputs to more sophisticated Gaussian mixture modeling \cite{TorrJOSA}.

 Here, based on the  local regression kernel center/surround features, the evidence from a location $\mathbf{r}$ of the frame is computed as   $P( \mathbf{I}^{(k)}(t) \mid  \mathbf{x}_{f}^{(k)}(\mathbf{r}, t) = 1, \mathbf{F}_{|\mathbf{I}}, \mathbf{r}_{F}(t)) = \frac{1}{\sum_{s}} \exp\left( \frac{ 1 - \rho(\mathbf{F}_{\mathbf{r}^{(k)},c}, \mathbf{F}_{\mathbf{r}^{(k)},s})}{\sigma^2} \right)$, where $\rho(\cdot) \in \left[-1,1 \right]$ is the matrix cosine similarity (see \cite{seo2009}, for details) between center and surround feature matrices $\mathbf{F}_{\mathbf{r}^{(k)},c}$ and $ \mathbf{F}_{\mathbf{r}^{(k)},s}$ computed at location $\mathbf{r}^{(k)}$   of the foveated frame. 

\begin{figure}[t!]
 \setlength{\tabcolsep}{2.1pt}
 \begin{center}
  \scriptsize
    \begin{tabular}{ccccc}
Input&Priority map&Proto objects&Interest points&\\
\includegraphics[width=0.09\textwidth]{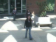}&
\includegraphics[width=0.09\textwidth]{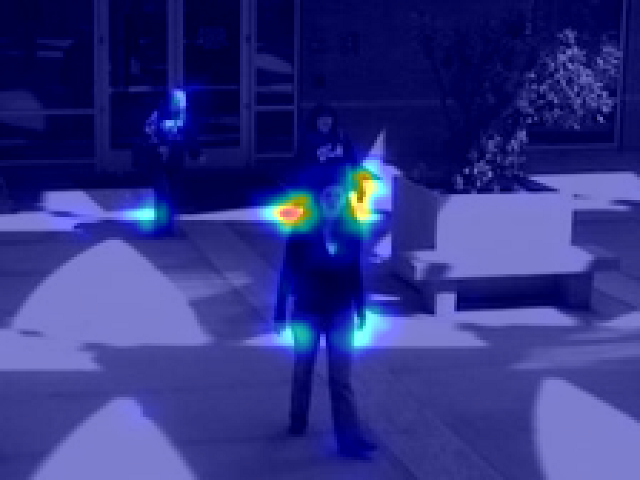}&
\includegraphics[width=0.09\textwidth]{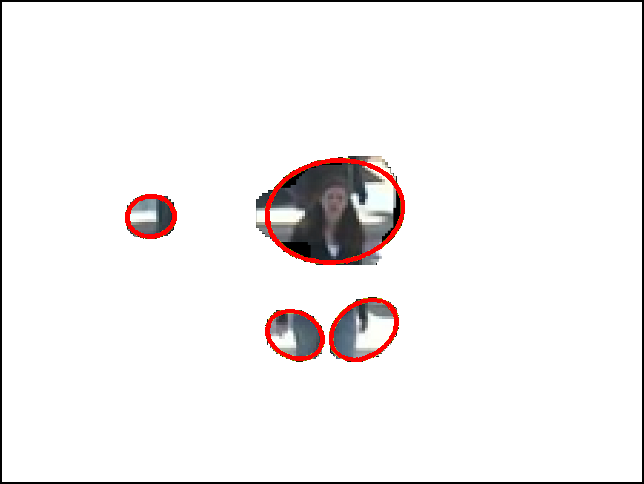}&
\includegraphics[width=0.09\textwidth]{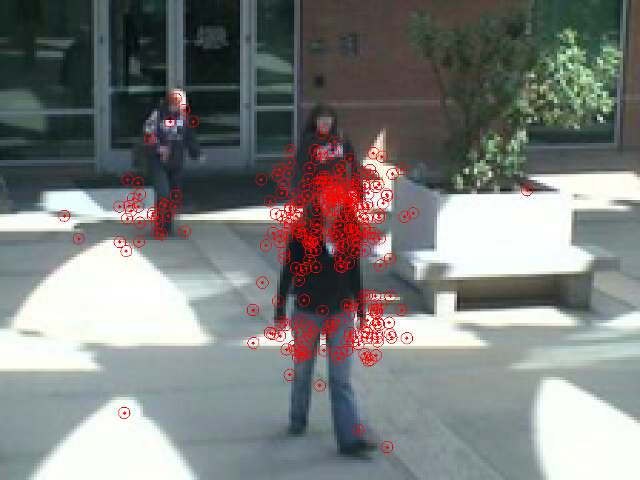}&\\
\\
\\
Input&Priority map&Proto objects&Interest points&Chosen FoA\\
\includegraphics[width=0.09\textwidth]{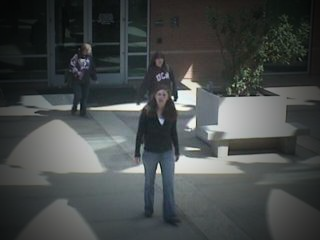}&
\includegraphics[width=0.09\textwidth]{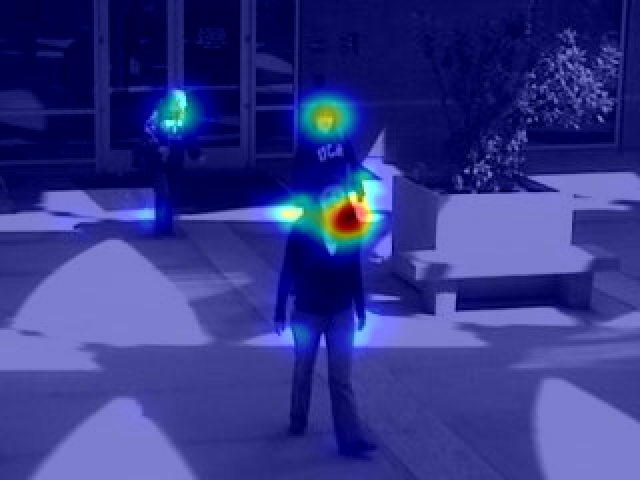}&
\includegraphics[width=0.09\textwidth]{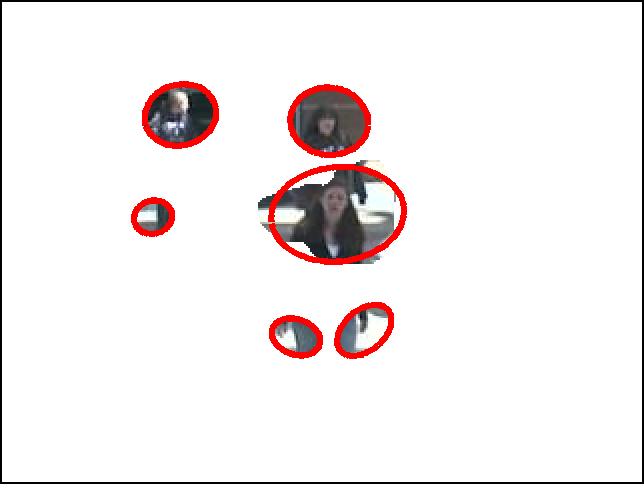}&
\includegraphics[width=0.09\textwidth]{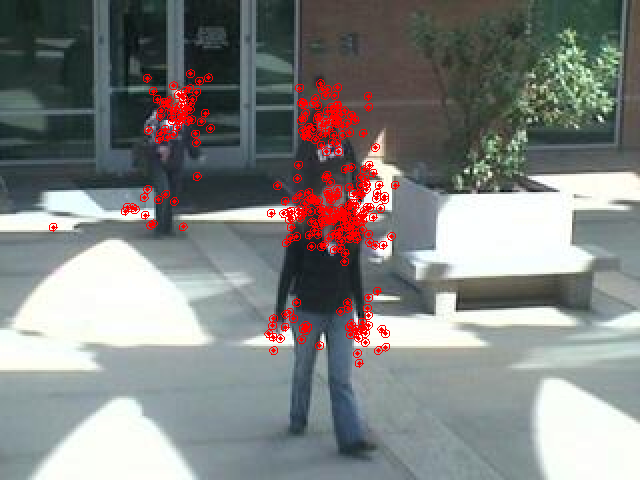}&
\includegraphics[width=0.09\textwidth]{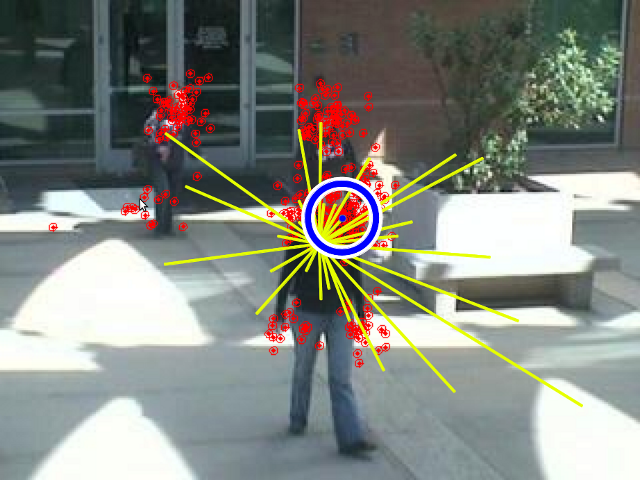}\\
    \end{tabular}
\caption{The main perceptual representation levels involved by  pre-attentive (top row) and attentive stages (bottom row). The input of the pre-attentive stage is the stream at low resolution. The priority map is visualized as a color map: reddish values specify most salient regions. Selected proto-objects are parametrised as ellipses.  IPs sampled from proto-objects are displayed as red dots (cfr. Section \ref{sec:landsampling}). The input of the attentive stage is the foveated stream  obtained by setting the initial FoA at the centre of the image. Candidates gaze shifts are displayed as yellow trajectories from the center of current FoA. The next FoA $\mathbf{r}_{F}(t+1)$ is chosen to maximise the expected reward (cfr. Section \ref{sec:patchsampling}), and displayed as a white/blue circle. 
}
  \label{fig:content_maps}
    \end{center}
\end{figure}

Figure~\ref{fig:content_maps} illustrates main representations discussed above (spatio-temporal priority maps,  proto-objects and   IPs sampled from proto-objects) 

\subsection{Action control}
\label{sec:action_contr}

The model exploits a  coarse-to-fine strategy. 
   First,  evaluation of   stream ``quality''  is pre-attentively performed, resorting to  the configurational complexity $\mathcal{C}^{(k)}(t)$ (cfr., Section~\ref{sec:landsampling}). 
This stage corresponds to  the  \emph{pre-attentive  loop} briefly summarised in Algorithm \ref{alg:pre}. 
\begin{algorithm}
\caption{Pre-attentive  loop}
\label{alg:pre}
\begin{algorithmic}[1]
\Statex \{\emph{Parallel execution on all streams} $1, \cdots, K$\} 
\Statex {\bf Input:}  $\{\mathbf{I}^{(k)}(t)\}_{k=1}^{K}$
\State Compute  bottom-up feature $\mathbf{F}^{(k)}_{|\mathbf{I}}$  and weight the feature map $\mathbf{X}_{f}^{(k)}(t)$.
\State Sample the priority map  $\mathbf{L}^{(k)}(t)$ conditioned on $\mathbf{X}^{(k)}(t)$.
\State Sample the potential object regions or proto-objects $\mathcal{O}^{(k)}(t)$ from $\mathbf{L}^{(k)}(t)$.
\State Based on available proto-objects $\mathcal{O}^{(k)}(t)$, sample IPs and compute the quality of the stream via  complexity $\mathcal{C}^{(k)}(t)$
\end{algorithmic}
\end{algorithm}
On this basis, the ``best'' quality stream is selected (cfr., Section~\ref{sec:patch_choice}),
and  the within-stream potential preys  are  attentively handled, in order to detect the actual  targets  $\mathbf{O}$ that are interesting under the given task $\mathbf{T}$ (cfr., Section~\ref{sec:patchsampling}).  Intra-stream behavior thus boils down to  an instance of the classic deployment of visual attention:  spotting an object  and keeping to it - via  either fixational movements or smooth pursuit -  or relocating to another  region (saccade) \cite{BocFerSMCB2013}. Thus, 
intra-stream behavior does not reduce  to  tracking, which  is to be considered solely the computational realization of visual smooth pursuit.  The attentive  loop is summarised in Algorithm \ref{alg:att}.
\begin{algorithm}
\caption{Attentive  loop}
\label{alg:att}
\begin{algorithmic}[1]
\Statex {\bf Input:}  $\{\mathbf{I}^{(k)}(t)\}_{k=1}^{K}$,  $\{\mathcal{C}^{(k)}(t)\}_{k=1}^{K}$
\Statex \{\emph{Patch choice}\}
\State Based on the  complexities $\{\mathcal{C}^{(k)}(t)\}_{k=1}^{K}$, sample the video stream $\widehat{k}$ to be analyzed.
\State Execute the \emph{between-stream gaze shift} $\mathbf{r}_{F}(t-1) \mapsto \mathbf{r}_{F}(t)$ at the center of current frame of stream $\widehat{k}$ and set the current FoA;
\Statex \Repeat
\State Compute bottom-up and top-down features $\{ \mathbf{F}^{(\widehat{k})}_{|\mathbf{I}}(t),\mathbf{F}^{(\widehat{k})}_{|\mathbf{O}}(t) \}$    and sample the feature map $\mathbf{X}^{(\widehat{k})}(t)$, based on $\mathbf{F}^{(\widehat{k})}_{|\mathbf{O}}(t)$
\State Sample the priority map  $\mathbf{L}^{(k)}(t)$ conditioned on $\mathbf{X}^{(k)}(t)$.
\State Sample proto-objects $\mathcal{O}^{(\widehat{k})}(t)$ from $\mathbf{L}^{(k)}(t)$;
\State Based on  $\mathcal{O}^{(k)}(t)$, sample IPs and compute the quality of the stream via  complexity $\mathcal{C}^{(\widehat{k})}(t)$
\Statex \{\emph{Prey handling}\}
\State Execute the \emph{within-stream  gaze shift} $\mathbf{r}_{F}(t) \mapsto \mathbf{r}_{F}(t+1)$ in order to maximize the expected reward with respect to the IP value and analyze the current FoA.
\Statex \Until {giving-up condition is met}
\Statex \{\emph{Patch leave}\}
\end{algorithmic}
\end{algorithm}

Clearly, since the number of targets is a priori unknown (partial information condition), efficient search requires tailoring a stopping decision to  target handling within the stream. 
 From a foraging standpoint, the decision to leave should also depend on future prospects for food on the current patch, which in turn depends on posterior information about this patch. 
 This issue is addressed in the framework of optimal Bayesian foraging~\cite{mcnamara2006bayes,killeen1996bayesian} (cfr., Section~\ref{sec:patchleave}).

 \section{Pre-attentive sensing}
 \label{sec:landsampling}

The goal of this stage is to infer a proto-object representation of all the $K$ patches within the spatial landscape (Fig.~\ref{fig:content_maps}, top row).
 To this end, the posterior 
$P(\mathbf{L}^{(k)}(t) \mid \mathbf{I}^{(k)}(t)) \approx P(\mathbf{L}^{(k)}(t),  \mathbf{I}^{(k)}(t)) $ 
is calculated from the joint pdf. In the  derivations that follows we omit the time index $t$ for notational simplicity.

Rewrite the joint pdf factorization in Eq. \ref{eq:joint} under the assumption of object-based feature independence, i.e., $ \prod_{f,f^{'}}  P(\mathbf{O},\mathbf{F}_{f |\mathbf{O}}^{(k)},  \mathbf{L}^{(k)}, \mathcal{O}^{(k)}, \mathbf{X}_{f}^{(k)}, \mathbf{F}_{f^{'}|\mathbf{I}}, \mathbf{I}^{(k)}  |  \mathbf{T}, \mathbf{r}_{F})$. 
Then $P(\mathbf{L}^{(k)},  \mathbf{I}^{(k)})$ is obtained by marginalizing over RVs $\mathbf{X}_{f}^{(k)}, \mathbf{F}_{f |\mathbf{O}}^{(k)}, \mathbf{F}_{f^{'} |\mathbf{I}}^{(k)},\mathbf{O}$ and $ \mathcal{O}^{(k)}$. Use the following: $\sum_{\mathcal{O}} P(\mathcal{O}^{(k)} | \mathbf{L}^{(k)}) =1$ by  definition; $P(\mathbf{L}^{(k)}) = Unif$ by assumption; $ \sum_{\mathcal{O}} P( \mathbf{F}_{f |\mathbf{O}}^{(k)} | \mathbf{O}) P(\mathbf{O} | \mathbf{T}) = P( \mathbf{F}_{f |\mathbf{O}}^{(k)} | \mathbf{T}) = P( \mathbf{F}_{f |\mathbf{O}}^{(k)})$ by local conditional independence in $\mathcal{G}$. Thus:
\begin{multline}
P(\mathbf{L}^{(k)} \mid \mathbf{I}^{(k)}) \approx 
 \prod_{f,f^{'}} \sum_{\mathbf{X}_{f}^{(k)},\mathbf{F}_{f |\mathbf{O}}^{(k)},\mathbf{F}_{f^{'} |\mathbf{I}}^{(k)}} P(\mathbf{X}_{f}^{(k)} \mid \mathbf{L}^{(k)}(t),\mathbf{F}_{f |\mathbf{O}}^{(k)}) \\
 P(\mathbf{I}^{(k)} \mid  \mathbf{F}_{f^{'} |\mathbf{I}}^{(k)},\mathbf{X}_{f}^{(k)},\mathbf{r}_{F}) P(\mathbf{F}_{f |\mathbf{O}}^{(k)}).
\label{eq:L_post}
\end{multline}
The  term $P(\mathbf{F}_{f |\mathbf{O}}^{(k)})$ is a prior ``tuning'' the preference for specific object-based features.
 In the pre-attentive stage we assume a uniform prior,  i.e. $P(\mathbf{F}_{f |\mathbf{O}}^{(k)})=Unif.$, and restrict to feed-forward features ${\mathbf{F}^{(k)}}_{|\mathbf{I}}$. Then, Eq. \ref{eq:L_post} boils down to the probabilistic form of a classic feed-forward saliency map (see Fig.~\ref{fig:content_maps}), namely, 
\begin{multline}
P(\mathbf{L}^{(k)} \mid \mathbf{I}^{(k)}) \approx 
\prod_{f,f^{'}} \sum_{\mathbf{X}_{f}^{(k)},\mathbf{F}_{f |\mathbf{O}}^{(k)},\mathbf{F}_{f^{'} |\mathbf{I}}^{(k)}} P(\mathbf{X}_{f}^{(k)} \mid \mathbf{L}^{(k)},\mathbf{F}_{f |\mathbf{O}}^{(k)}) \cdot  \\P(\mathbf{I}^{(k)} \mid \mathbf{F}_{f^{'} | \mathbf{I}}^{(k)}, \mathbf{X}_{f}^{(k)},\mathbf{r}_{F}),
\label{eq:L_sal}
\end{multline}
\noindent 
\noindent where the likelihood $P(\mathbf{X}_{f}^{(k)} |\mathbf{L}^{(k)},\mathbf{F}_{f |\mathbf{O}}^{(k)})$ is modulated by bottom-up feature likelihood $P(\mathbf{I}^{(k)} | \mathbf{F}_{f^{'} |\mathbf{I}}^{(k)}, \mathbf{X}_{f}^{(k)},\mathbf{r}_{F})$.

Given the priority map, a set $\mathcal{O}^{(k)}(t) =\{\mathcal{O}^{(k)}_p(t)\}_{p=1}^{N_P}$ of  $N_P$ proto-objects or candidate preys  can be sampled from it. Following \cite{BocFerSMCB2013}, we  exploit  a sparse representation of proto-objects. These are conceived   in terms of  ``potential bites", namely interest points   sampled from the proto-object. 
At any given time $t$,  each proto-object is characterised by different shape and location, i.e.,   $\mathcal{O}^{(k)}_p(t)=(O^{(k)}_p(t), \Theta_p(t))$. 
Here  $O^{(k)}_p(t) =  \{ \mathbf{r}^{(k)}_{i,p}\}_{i=1}^{N_{i,p}}$ is the sparse representation of proto-object $p$ as the cluster of $N_{i,p}$ IPs   sampled from it;  $\Theta^{(k)}_p(t)$ is a parametric  description of a proto-object,  $\Theta^{(k)}_p(t)= (\mathcal{M}^{(k)}_p(t), \theta^{(k)}_p)$.

The set $\mathcal{M}^{(k)}_p(t)=\{m^{(k)}_p(\mathbf{r}, t)\}_{\mathbf{r} \in L}$ stands for a map of binary RVs indicating at time $t$  the presence or absence of proto-object $p$, and the overall map of proto-objects  is given by $\mathcal{M}^{(k)}(t) = \bigcup _{p=1}^{N_p} \mathcal{M}^{(k)}_p(t)$. Location and shape of the proto-object are parametrized via  $\theta^{(k)}_p$.   Assume  independent proto-objects: 
\begin{equation} 
\mathcal{M}^{(k)}(t) \sim P( \mathcal{M}^{(k)}(t) | \mathbf{L}^{(k)}(t)),
\label{eq:sample_M}
\end{equation}
and for $p=1,\cdots,N_P$
\begin{equation} 
\theta^{(k)}_p(t) \sim P( \theta^{(k)}_p(t) | \mathcal{M}^{(k)}_{p}(t)=1, \mathbf{L}^{(k)}(t)), 
\label{eq:sample_theta}
\end{equation}
\begin{equation}
O^{(k)}_p(t) \sim P( O^{(k)}_p(t)| \theta^{(k)}_p(t), \mathcal{M}^{(k)}_p(t)=1, \mathbf{L}^{(k)}(t)).
\label{eq:sample_O}
\end{equation}
The first step (Eq. \ref{eq:sample_M})  samples the proto-object map from the landscape. The second (Eq. \ref{eq:sample_theta}) samples proto-object parameters $ \theta(t)^{(k)}_p=(\mu^{(k)}_p(t), \Sigma^{(k)}_p(t)))$.

Here, $\mathcal{M}^{(k)}(t)$ is  drawn from the priority map  by deriving a preliminary binary map $\widetilde{\mathcal{M}}^{(k)}(t) = \widehat{m}^{(k)}(\mathbf{r}, t)\}_{\mathbf{r} \in L}$, such that $\widehat{m}^{(k)}(\mathbf{r}, t) =1$ if $P(\mathbf{L}^{(k)}(t) | \mathbf{I}^{(k)}(t)) > T_M$, and $\widehat{m}^{(k)}(\mathbf{r}, t) =0$ otherwise. The  threshold $T_M$ is  adaptively set so as to achieve  $95$\% significance level in deciding whether the given priority values are in the
extreme  tails of the  pdf. The procedure is based on the assumption that  an informative proto-object is a relatively rare region and thus results in values which are in the tails of $P(\mathbf{L}^{(k)}(t) | \mathbf{I}^{(k)}(t))$. Then,  following \cite{walther2006},    $\mathcal{M}^{(k)}(t)=\{\mathcal{M}^{(k)}_p(t)\}_{p=1}^{N_P}$  is obtained as 
$\mathcal{M}^{(k)}_p(t)=\{m^{(k)}_p(\mathbf{r}, t) |   \ell(B, \mathbf{r}, t)=p \}_{\mathbf{r} \in L}$, where the function $\ell$  labels $\widetilde{\mathcal{M}}(t)$ around $\mathbf{r}$. 

 We set the maximum number of proto-object to $N_P = 15$ to retain the most important ones. 
 
As to Eq. \ref{eq:sample_theta},  the proto-object map provides the necessary spatial support  for a 2D ellipse maximum-likelihood approximation of each proto-object, whose location and shape are parametrized as $ \theta^{(k)}_p=(\mu^{(k)}_p, \Sigma^{(k)}_p)$ for $p=1,\cdots,N_p$    (see \cite{BocFerSMCB2013} for a formal justification).
 
In the third step (Eq. \ref{eq:sample_O}), the procedure generates   clusters  of IPs, one cluster for each proto-object $p$ (see Fig. \ref{fig:content_maps}). By assuming a Gaussian distribution centered on the proto-object - thus with mean $\mu^{(k)}_p$ and covariance matrix  $ \Sigma^{(k)}_p$ given by the axes parameters of the 2D ellipse fitting the proto-object shape -, 
Eq. (\ref{eq:sample_O}) can be further specified  as \cite{BocFerSMCB2013}:
\begin{equation} 
\mathbf{r}^{(k)}_{i,p} \sim  \mathcal{N}(\mathbf{r}^{(k)}_p; \mu^{(k)}_p(t), \Sigma^{(k)}_p(t)), i=1,\cdots, N_{i,p}.
\label{eq:sample_rip}
\end{equation} 

We set $N_s=50$ the maximum number of IPs and for each proto-object  $p$, we sample $ \{\mathbf{r}^{(k)}_{i,p}\}_{i=1}^{N_{i,p}}$ from a Gaussian centered on the proto-object as in (\ref{eq:sample_rip}). The number of IPs per proto-object is estimated as $N_{i,p}= \lceil N_s \times \frac{A_p}{\sum_p A_p }\rceil$, $A_p=\pi \sigma_{x,p}\sigma_{y,p} $ being the size (area) of proto-object $p$.
Eventually, the set of all IPs characterising the pre-attentively perceived proto-object can be obtained as $O(t) = \bigcup _{p=1}^{N_p} \{ \mathbf{r}^{(k)}_{i,p}(t)\}_{i=1}^{N_{i,p}}$.

\section{Stream selection}
\label{sec:patch_choice}
Streams vary in the number of objects they contain and maybe other characteristics such as the ease with which individual items are found. We assume that in the pre-attentive stage, the choice of the observer to spot  a stream, is drawn on the basis of some global index of interest characterizing each stream in the visual landscape.
In ecological modelling  for instance,    one such index is the landscape entropy  determined by dispersion/concentration of  preys \cite{stephens1986foraging}. 
 
Here, generalizing these assumptions,  we introduce the time-varying configurational complexity $\mathcal{C}^{(k)}(t)$ of  the  $k$-th stream.
Intuitively, by considering each stream a dynamic system, we resort to the general principle that  
complex systems are neither completely  random neither perfectly ordered and complexity should reach its maximum at a level of  randomness away from these extremes~\cite{shiner1999}. 
For instance,  a crowded scene with many pedestrians moving represents a disordered system (high entropy, low order) as opposed to a  scene where no activities take place (low entropy, high order). The highest complexity is thus reached when specific activities occur: e.g., a group of people meeting.   To formalize the relationship between  stream complexity  and stream selection    we proceed as follows. Given $\mathcal{C}^{(k)}(t), k=1,\cdots,K$, 
the choice of the $k$-th stream is obtained by sampling from the categorical distribution 
\begin{equation}
k \sim \prod_{k=1}^{K} \left[P(\mathcal{C}^{(k)}(t))\right]^{k},
\label{eq:patch_sample}
\end{equation} 
\noindent with
 \begin{equation}
P(\mathcal{C}^{(k)}(t)) = \frac{\mathcal{C}^{(k)}(t))}{\sum_{k=1}^{K} \mathcal{C}^{(k)}(t))}.
\label{eq:P_C}
\end{equation} 
\noindent Keeping to~\cite{shiner1999},   complexity  $\mathcal{C}^{(k)}(t)$ is defined in terms of order/disorder of the system,
 \begin{equation}
 \mathcal{C}^{(k)}(t)= \Delta^{(k)}(t) \cdot \Omega^{(k)}(t),
  \label{eq:compl}
  \end{equation}
 \noindent where $\Delta^{(k)} \equiv H^{(k)}/H^{(k)}_{sup}$   is the disorder parameter,  $\Omega^{(k)} = 1- \Delta^{(k)}$  is the order
parameter, and  $H^{(k)}$ the Boltzmann-Gibbs-Shannon (BGS) entropy  with $H^{(k)}_{sup}$ its  supremum. $H^{(k)}$ and $H^{(k)}_{sup}$ are calculated as follows.

For each stream $k$, we compute the BGS entropy $H$ as a function of  the spatial  configuration of the sampled IPs. The spatial domain $\mathcal{D}$  is partitioned into a configuration space of cells (rectangular windows), i.e.,  $\{w(\mathbf{r}_c)\}_{c=1}^{N_w}$, each cell being centered at  $\mathbf{r}_c$. By assigning  each IP  to the corresponding window, the probability  for point $\mathbf{r}_{s}$  to be within  cell $c$ at time $t$ can be estimated   as 
$P^{(k)}(c,t)  \simeq \frac{1}{N_s}\sum_{s=1} ^{N_s}\chi_{s,c}$,  
where $\chi_{s,c}=1$ if  $\mathbf{r}_{s} \in w(\mathbf{r}_c) $
and $0$ otherwise.

Thus,  $H^{(k)}(t)=-k_B\sum_{c=1}^{N_w} P^{(k)}( c,t )  \log P^{(k)}(c,t) $, and (\ref{eq:compl}) can be easily computed. Since  dealing with a fictitious thermodynamical system, we set Boltzmann's constant $k_B=1$. The supremum of $H^{(k)}(t)$ is  $H_{sup}=\log N_w$ and it is associated to a completely unconstrained process, that is a process where $H^{(k)}(t)=const$, since with reflecting boundary conditions the asymptotic  distribution is uniform. 

When stream $k$ is chosen at time $t-1$, attention is deployed to the stream via the gaze shift $\mathbf{r}_{F}(t-1) \rightarrow \mathbf{r}_{F}(t)$, and the ``entering time'' $t_{in} = t$ is set.

\section{Attentive stream handling}
\label{sec:patchsampling}

When gaze is deployed to the $k$-th stream,  the $\mathbf{r}_{F}(t_{in})$ is positioned at the centre of the frame, and  foveation  is  simulated by blurring  $\mathbf{I}^{(k)}(t_{in})$ through  an isotropic Gaussian function centered at $\mathbf{r}_{F}(t_{in})$,  whose variance is taken as the radius of a FoA, $\sigma=|FOA|$. This is  approximately given by $1/8 \min [width, height]$, where $width\times height = |\mathcal{D}|$, $|\mathcal{D}|$ being the dimension of the frame support $\mathcal{D}$. This way we obtain the  foveated image,  which provides  the input for the next processing steps. The foveation process  is updated for  every gaze shift within the patch that involves a large relocation (saccade), but not during small relocations, i.e.  fixational or pursuit eye movements.
At this stage, differently from pre-attentive analysis, the observer  exploits the full priority posterior as formulated in Eq.~\ref{eq:L_post}, rather than the reduced form specified in Eq.~\ref{eq:L_sal}. In other terms, the object-based feature likelihood, $P( \mathbf{F}^{(k)}_{|\mathbf{O}}|\mathbf{O})$, is taken into account.

Object search  is performed by sampling, from current location $\mathbf{r}_{F}$,  a set of candidate gaze shifts  $\mathbf{r}_{F}(t) \rightarrow \mathbf{r}^{(k)}_{new}(t+1)$ (cfr. Fig.\ref{fig:content_maps}, bottom-right picture). In  simulation, candidate point sampling is performed as in  \cite{BocFerSMCB2013}. In a nutshell, $\mathbf{r}^{(k)}_{new}(t+1)$ are sampled via a Langevin-type stochastic differential equation, where the drift component is a function of IPs' configuration, and the stochastic component is sampled from the  L\'evy $\alpha$-stable distribution. The latter accounts for prior oculomotor biases on gaze shifts. We use different $\alpha$-stable parameters for the different types of gaze shifts - fixational, pursuit and saccadic shifts -, that have been learned from eye-tracking experiments of human subjects observing videos under the same task considered here. The time-varying choice of the family of parameters is conditioned on the current complexity index $\mathcal{C}^{(k)}(t)$ (\cite{BocFerSMCB2013} for details).

Denote $R^{(k)}$  the reward consequent on a gaze shift.  Then,  next  location is chosen  to maximize the expected reward:
\begin{equation}
\mathbf{r}_{F}(t+1) = \arg\max_{\mathbf{r}^{(k)}_{new}} E\left[R^{(k)}_{\mathbf{r}^{(k)}_{new}}\right] .
\label{eq:maxrew}
\end{equation}
The expected reward is computed with reference to the  value  of  proto-objects available within the stream,
\begin{equation}
E\left[R^{(k)}_{\mathbf{r}^{(k)}_{new}}\right] = \sum_{p \in \mathcal{I}^{(k)}_{V} } Val(\mathcal{O}^{(k)}_{p}(t)) P(\mathcal{O}^{(k)}_{p}(t) | \mathbf{r}^{(k)}_{new}(t+1),\mathbf{T}).
\label{eq:exprew}
\end{equation}

\noindent Here  $Val$ is the average value of proto-object $\mathcal{O}_{p}(t)$ with respect to  the posterior $P(\mathbf{L}^{(k)}(t) | \mathbf{I}^{(k)}(t))$, which, by using samples generated via Eq. \ref{eq:sample_rip}, can be simply evaluated as
\begin{equation}
Val(\mathcal{O}^{(k)}_{p}(t)) \simeq  \sum_{i \in \mathcal{I}_{p}}  P(L_{i}^{(k)}(t) | \mathbf{I}^{(k)}(t)).
\label{eq:value}
\end{equation}

The observer samples $N_{new}$  candidate gaze shifts. Using  Eqs. \ref{eq:sample_rip} and \ref{eq:value},  Eq. \ref{eq:exprew} can be written as 
\begin{multline}
E\left[R^{(k)}_{\mathbf{r}^{(k)}_{new}}\right] =\\ \sum_{p \in \mathcal{I}^{(k)}_{V} }  \sum_{i \in \mathcal{I}_{p}} Val(\mathbf{r}^{(k)}_{i,p}(t)) \mathcal{N}(\mathbf{r}^{(k)}_{i,p}(t) | \mathbf{r}^{(k)}_{new}(t+1),\Sigma_s),
\label{eq:exprew2}
\end{multline}
\noindent where $\Sigma_s$ defines the  region around $\mathbf{r}^{(k)}_{new}(t+1)$. In foraging terms, Eq. \ref{eq:exprew} formalises the  expected reward of gaining valuable bites of food (IPs) in the neighbourhood of the candidate shift $\mathbf{r}_{new}$.

Note that  effective reward $R^{(k)}(t)$ is gained by the observer only if the gaze shift is deployed to a point $\mathbf{r}$ that sets a FoA overlapping an object of interest for the task (in the simulation, for simplicity, $R^{(k)}(t)=1$  when a face or a body is detected, and $0$ in other cases). Thus, as the observer attentively explores the stream, he updates his estimate of stream quality in terms of accumulated rewards, which will provide the underlying support  for the stream giving-up strategy.

A final remark concerns the number of objects that can be detected  within the stream. Attentive analysis is sequential by definition. In principle, all relevant objects in the scene can be eventually scrutinized, provided that enough time is granted to the observer. For instance, as to detection performance, current implementation of the model exploits \emph{adaboost} face and body detectors that have been trained on a much larger dataset than original Viola-Jones detectors, leading to about  $90\%$ detection accuracy (considering a minimal detectable region of  $40 \times 40$ pixel area).   But cogently,   the actual number of scrutinized objects is the result of  observer's  trade-off between the quality of the visited stream   and the potential quality of the other $K-1$ streams. Namely, it depends on the stream giving-up time as dynamically determined by the Bayesian strategy.

\section{The Bayesian giving-up strategy}
\label{sec:patchleave}
In this Section we consider the core problem of switching from one stream  to another. In foraging theory  this issue is addressed as ``How long should a forager persevere in a patch?''. Two approaches can be pursued: i) patch-based or global/distal models; ii) prey-based or local/proximal models. These are, for historical reasons,  subject to separate analyses and modeling \cite{stephens1986foraging}.  The Bayesian strategy we propose here aims at filling such gap.   

\subsection{Global models. Charnov's Marginal Value Theorem}
In the  scenario envisaged by Charnov~\cite{charnov1976optimal} the landscape is composed of food patches that deliver food rewards as a smooth decreasing flow.  Briefly, Charnov's MVT states that a patch leave decision should be taken when the expected current rate of information return falls below the mean rate that can be gained from other patches. MVT considers food intake as a continuous deterministic process where foragers assess patch profitability by the instantaneous net energy intake rate. In its original formulation, it provides the optimal solution to the problem, although only once the prey distribution has already been learnt;  it assumes omniscient foragers (i.e. with a full knowledge of preys and patch distribution).  The model is purely functional,  nevertheless  it is important for generating two testable qualitative predictions~\cite{vanAlphen2008}: 1) patch time should increase with prey density in the patch; 2) patch times should increase with increasing average travel time in the habitat and should decrease with increasing average host density in the patches.

\subsection{Local models}

The MVT and its stochastic generalization do not take into account the behavioral \emph{proximate} mechanisms used by foragers to control patch time or to obtain information about prey distribution \cite{vanAlphen2008}.
Such a representation of intake dynamics is inadequate to account for the real search/capture processes occurring within the patch. These, in most cases, are discrete and stochastic events in nature.   
For instance, Wolfe  \cite{wolfe2013time} has examined human foraging in a visual search context,  showing that  departures from MVT emerge when patch quality varies and when visual information is degraded. 

Experience on a patch, in terms of cumulative reward,  gives information on current patch type and on future rewards.  A good policy should make use of this information and vary the giving-up time with experience. In this perspective, as an alternative to MVT,  local models, e.g., Waage's   \cite{waage1979foraging}, assume that the motivation of a forager to remain and search on a particular patch  would be linearly correlated with  host density. As long as this ``responsiveness'' is above a given (local) threshold, the forager does not leave the patch~\cite{waage1979foraging}. As a consequence, the total time spent within the patch, say $\Delta^{(k)}_w$, eventually depends on the experience of the animal within that patch.

\subsection{Distal and proximal strategies in an uncertain world}

To deal with uncertainty \cite{mcnamara2006bayes,killeen1996bayesian}, 
 a forager should persevere in a patch as long as the probability of the next observation being successful is greater than the probability of the first observation  in one among the $K-1$ patches being successful, taking into account the time it takes to make those observations.

Recall that  complexity $\mathcal{C}^{(k)}(t)$  is used as a pre-attentive stochastic proxy of the likelihood that the $k$-th stream yields a reward $R^{(k)}(t)$ to the observer. 
Thus, $P(\mathcal{C}^{(k)}(t))$ defined in Eq. \ref{eq:P_C} stands for the prior probability of objects being primed for patch $k$ (in OFT,  the base rate \cite{stephens1986foraging}).

A common detection or gain function, that is the probability of reinforcement vs. time, is an exponential distribution of times to detection \cite{killeen1996bayesian}, and can be defined in terms of the conditional probability of gaining a reward in stream $k$ by time $t$, given that it has been primed via complexity  $\mathcal{C}^{(k)}(t)$:
\begin{equation}
P(R^{(k)}(t) \mid \mathcal{C}^{(k)}(t))  = 1- \exp(-\lambda t)
\label{eq:gain}
\end{equation} 
\noindent where $\lambda$ is the detection rate. Then, by generalising the two patch analysis discussed in \cite{killeen1996bayesian},   the following holds.

\begin{proposition}
\label{prop:strategy}
Denote $ \langle \mathcal{C}^{(k)}(t) \rangle_{i \neq k}$ the average complexity of the $K-1$ streams other than $k$. Under the hypothesis that at $t=0$, ${C}^{(k)}(0) >  \langle \mathcal{C}^{(k)}(t) \rangle_{i \neq k}$,  leaving the $k$-th stream when 
\begin{equation}
\mathcal{C}^{(k)}(t) \exp(-\lambda t)  =   \langle \mathcal{C}^{(k)}(t) \rangle_{i \neq k}, t > 0,
\label{eq:opt}
\end{equation}
defines an optimal Bayesian strategy for the observer.
\end{proposition}
\begin{IEEEproof}
See Appendix \ref{app:opt}.
\end{IEEEproof}

The strategy summarised via Eq.~\ref{eq:opt} can  be considered as a Bayesian version of the MVT-based strategy ~\cite{charnov1976optimal}.
 In order to reconcile the \emph{distal} functional constraint formalised through  Eq. \ref{eq:opt}, with the behavioral proximate mechanisms used by foragers within the patch, we put a prior distribution on the $\lambda$ parameter of the exponential distribution in the form of a Gamma distribution, i.e., $Gamma (\lambda; \nu^{(k)}, \Delta^{(k)})$, where $\nu^{(k)}, \Delta^{(k)}$ are now hyper-parameters governing the distribution of the $\lambda$ parameter.  Assume that when the observer selects the stream, the initial prior is  $Gamma (\lambda; \nu^{(k)}_0, \Delta^{(k)}_0)$.  

The hyper-parameters $\nu^{(k)}_0, \Delta^{(k)}_0$ represent initial values of  expected rewards and   ``capture'' time, respectively, thus  stand for  ``a priori'' estimates of stream profitability. For $t > t_{in}$, the posterior over $\lambda$ can be computed via Bayes' rule as $Gamma (\lambda; \nu^{(k)}, \Delta^{(k)}) \propto \exp(-\lambda t) Gamma (\lambda; \nu^{(k)}_0, \Delta^{(k)}_0)$. Since the Gamma distribution is a conjugate prior, the  Bayesian update  only calls for the determination of the hyper-parameter  update 

\begin{equation}
\nu^{(k)} =  \nu^{(k)}_0 + n,  \;  \; \;  \; \;  \; \;  \;    \Delta^{(k)} =   \Delta^{(k)}_0 + \sum_{i=1}^{n}  \Delta(t_{n}),
\label{eq:k}
\end{equation}

\noindent $n$ being the number of handled objects, that is the number of rewards effectively gained up to current time, i.e., $\sum_{t=t_{in}}^{t'}R^{(k)}(t)$, and $\Delta(t_{n})$ the interval of time spent on the $n$-th proto-objects. The latter, in general,  can be further decomposed as $\Delta(t_{n}) = TD_n + TH_n$, where $TD_n$ and $TH_n$ denote the time to spot and handle the $n$-th proto-object, respectively. Clearly, time $TD_n$ elapses for any proto-object within the stream, whilst  $TH_n$ is only taken into account  when the object has been detected as such (e.g., a moving proto-object as a pedestrian) and actual object handling occurs (e.g., tracking the pedestrian), otherwise $TH_n = 0$. In the experimental analyses we will assume, for generality, $TD_n = \delta_D \phi(| Proto |)$ and $TH_n = \delta_H \phi(| Object |)$,  where $\delta_D$ and $\delta_H$ are times to process elements (pixels, super-pixels, point representation  or parts) defining the prey,  which depend on the specific algorithm adopted;  $\phi(| \cdot |)$ is a function (linear, quadratic, etc) of the dimension of the processed item.

Eventually, when hyper-parameters  have been computed  (Eq. \ref{eq:k}), a suitable value for $\lambda$ can be obtained as the expected value $\overline{\lambda} = E_{Gamma}\left[ \lambda \right]= \frac{\nu^{(k)}}{\Delta^{(k)}} $. As a consequence, the total within-stream time $\Delta^{(k)}_w$  depends on the experience of the observer within that stream

Here, this proximal mechanism is formally related to the distal global quality of all streams, via the condition specified through Eq.~\ref{eq:opt} so that the decision threshold is dynamically modulated by the pre-attentive observer's perception across streams. As a result, even though on a short-time scale the observer might experience local motivational increments due to rewards, on a longer time scale the motivation to stay within the current stream  will progressively decrease.

\section{Experimental work}
\label{sec:exp}

\subsection{Dataset}

We used a portion of the the UCR Videoweb Activities Dataset~\cite{denina2011videoweb}, a publicly available dataset containing  data recorded from multiple outdoor wireless cameras. The dataset contains $4$ days  of recording and several scenes for each day,  about $2.5$ hours of video  displaying dozens of activities along with annotation. For the first three days, each scene is composed of a collection of human activities and motions which forms a continuous storyline. 

The dataset is designed for evaluating the performance of human-activity recognition algorithms, and it  features multiple human activities viewed from multiple cameras located asymmetrically with overlapping and non-overlapping views, with varying degrees of illumination and lighting conditions. This amounts to a large variety of simple actions  such as walking, running, and waving.  

We experimented on three different scenes recorded in three different days. Here we present results obtained from scene 1, recorded in the second day (eight camera recordings). Results  from the other scenes are reported as Supplementary Material. The scene contains the streams  identified by the following ids: $cam16$, $cam17$, $cam20$, $cam21$, $cam27$, $cam31$, $cam36$, $cam37$. Each video is at $30$ fps and cameras are not time-synchronized. We synchronized video streams by applying the following shifts between cameras: $[cam 16:291,cam 17: 191,cam 20: 0,cam 21: 0,cam 27: 389,cam 31: 241,cam 36: 0,cam 37: 373]$. Cameras $cam20$, $cam21$ and $cam36$ can be used as time reference.  Since the video of the camera $cam21$ is the shortest ($\approx 8000$ frames), the analyzes presented in the following  consider the frames between $1$ and $8000$.  

Annotated activities  are: \emph{argue within two feet}, \emph{pickup object}, \emph{raised arms}, \emph{reading book}, \emph{running}, \emph{sit cross legged}, \emph{sit on bench}, \emph{spin while talking}, \emph{stand up}, \emph{talk on phone}, \emph{text on phone}. All  are performed by humans.

As previously discussed, we are not concerned with action or activity recognition. Nevertheless,  the dataset provides a suitable benchmark.   The  baseline aim  of the model  is to dynamically set the FoA on the most informative subsets of the video streams  in order to capture  atomic events  that are at the core of the activities actually recorded.  In this perspective, the virtual forager operates under the task  ``pay attention to people within the scene'', so that the classes of objects of interest are represented by faces and human bodies. 
The output collection of subsets from all streams can eventually  be evaluated in terms of the retrieved activities marked in the ground-truth.

\subsection{Experimental evaluation}

Evaluation of results should  consider  the two  dimensions of  i) visual representation and  ii) giving-up strategy. For instance,  it is not  straightforwardly granted that  a pre-attentive representation for choosing the patch/video  might perform better (beyond computational efficiency considerations) than an attentive representation, where all objects of interest are detected before selecting the video stream. 

As to the giving-up time choice,  any strategy should in principle perform better than  random choice. Again, this should not be given for granted, since in  a complex scenario a bias-free, random allocation  could perform better than  expected. Further, a pure Charnov-based strategy, or a deterministic one, e.g. \cite{napoletano2014attentive}, could offer  a reasonable solution.  Under this rationale, evaluation takes into account the following analyses.

\subsubsection{Representations of visual information}
Aside from the basic priority map representation (denoted  M in the remainder of this Section),  which is exploited  by our model (Eqs.~\ref{eq:L_sal} and ~\ref{eq:L_post}  for the pre-attentive and attentive  stages, respectively), the following alternatives have been considered.  

\begin{itemize}
\item \emph{Static} (denoted S): the baseline salience computation  by Itti \emph{et al.}~\cite{itti1998model}. The method combines  orientation, intensity and color contrast features in a purely bottom-up scheme. The frame-based  saliency map is converted  in a probability map (as in \cite{bocc08tcsvt}) so to implement  the bottom-up priority map (Eq.~\ref{eq:L_sal}). Attentive exploration is driven by bottom-up information and object-based likelihood is kept uniform when computing Eq.~\ref{eq:L_post}.

\item \emph{Static combined with Change Detection and Face/Body Detection} (S+CD+FB): this representation has been used in \cite{napoletano2014attentive}.  It draws on the Bayesian integration of top-down / bottom-up information as  described in \cite{bocc08tcsvt}. Novelties are computed by detecting changes  between two subsequent frames  at a lower spatial resolution~\cite{napoletano2014attentive}. 
In our setting, it amounts to assume that the observer has the capability of detecting objects  before selecting the stream; namely, it boils down to directly compute Eq.~\ref{eq:L_post}.

\item \emph{Proposed model with early prey detection}  (M+): akin to the   S+CD+FB scheme, the full priority probability  (Eq.~\ref{eq:L_post}) is exploited before stream selection, instead of the bottom-up priority (Eq.~\ref{eq:L_sal}).  
\end{itemize}

Clearly, there are differences between adopting one representation or the other. These can be readily appreciated by analyzing  behavior over time of  stream complexities $\mathcal{C}^{(k)}, k=1,\cdots,K$  obtained by adopting the above representations.  One stream is hardly distinguishable from another when  using the S  and  the S+CD+FB representations; by contrast, higher discriminability is apparent for the M and M+ settings (cfr. Fig 12 and 13, Supplementary Material). Yet, most interesting here is to consider representational performance as related to  foraging strategy.

\subsubsection{Foraging strategies}
As to stream giving-up, we compare the following strategies.

\paragraph{Deterministic}
The simplest  strategy~\cite{stephens1986foraging}. A camera switch is triggered after a fixed time $\Delta_w >0$. Higher values of within-stream time $\Delta_w$ entail a low number of  switches.

\paragraph{Random}
This strategy triggers a camera switch after a random time $\Delta_w$. In this case $\Delta_w$ is a RV drawn from a uniform pdf $Unif(0,b_w)$, where $b_w$ is a suitable parameter. 

\paragraph{Charnov}
We adapted  the solution to  Charnov's MVT ~\cite{charnov1976optimal}  by Lundberg et \emph{al} ~\cite{lundberg1990functional}. If the observer chooses stream $k$ at time $t_{in}$, the optimal stream residence time  is defined as $\Delta^{(k)}_w = \mathcal{C}^{(k)}(t_{in}) \cdot \sqrt{\frac{t_b}{\langle \mathcal{C}(t_{in})\rangle \cdot \delta}}$,  where $\mathcal{C}^{(k)}(t_{in})$ is the resource level  in stream $k$ (here assessed in terms of complexity) at entering time $t_{in}$, $\langle \mathcal{C}(t_{in})\rangle$ is the average resource level across streams, $\delta$ is a parameter determining the initial slope of the gain function in the stream, and $t_b$ is the average switching (travelling) time between  two video streams. By assuming  constant traveling time  ($t_b =1$),  the only parameter to  determine is the slope $\delta>0$. Note that, when $\mathcal{C}^{(k)}(t_{in})>\langle \mathcal{C}(t_{in})\rangle$, higher values of $\mathcal{C}^{(k)}(t_{in})$  entail higher values of $\Delta^{(k)}_w$.

\subsection{Evaluation measures}

\begin{figure}[t]
  \centering
  \includegraphics[width=0.38\textwidth]{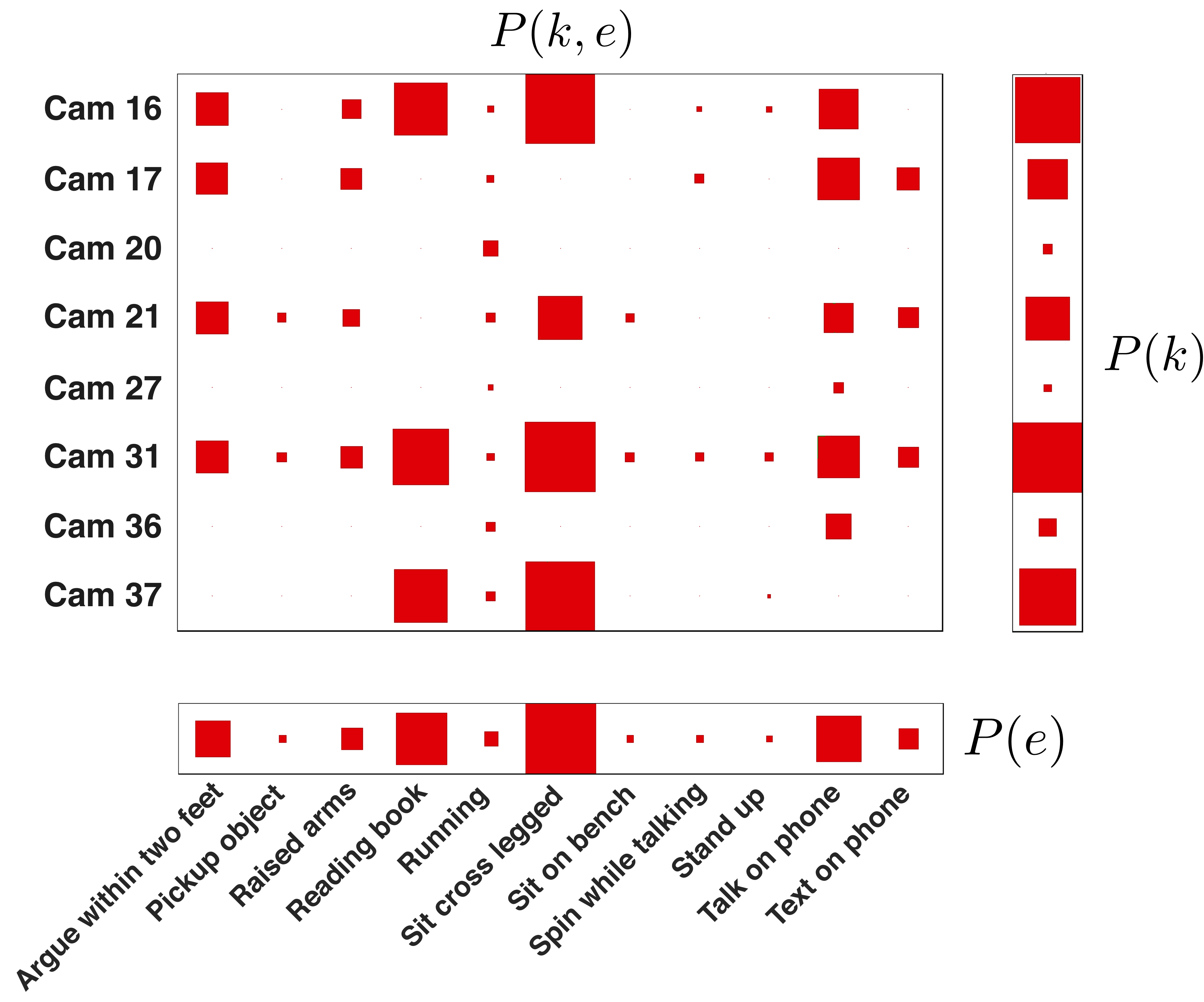}
    \caption{A probabilistic glance at the information content of multiple video streams:  the distribution  of activities  within each stream of the dataset represented in terms of joint and marginal probability distributions $P(k,e)$, $P(k)$ and $P(e)$, respectively. Distributions are visualised  as  Hinton diagrams, i.e., the square size is proportional to the probability value. The plots for the marginal distributions have different scales from those for the joint distribution (on the same scale, the marginals would look larger as they sum all of the mass from one direction).}
  \label{fig:events_distr}
\end{figure}

\begin{figure}[tb]
  \centering
   \includegraphics[width=0.34\textwidth]{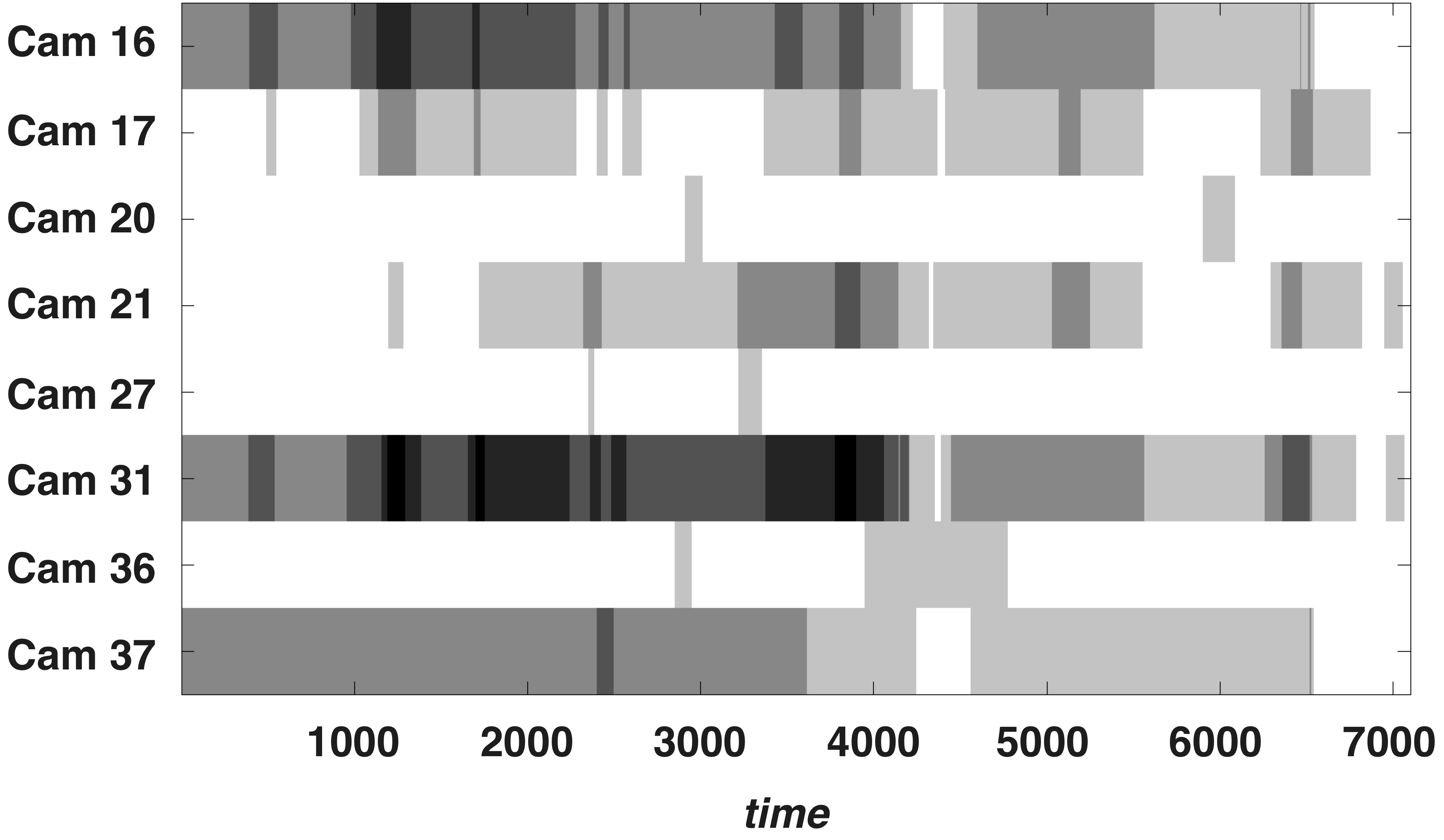}
  \caption{Distribution of activities across cameras and time. A darker color indicates that several activities co-occur at the same time while the white color indicates the total absence of activities.}
  \label{fig:gt}
\end{figure}

The definition of  measures that capture the subtleties of activity dynamics across multiple cameras is not straightforward. 
One has to account for the overall distribution of activities with respect to the different streams. Meanwhile, each  stream should be characterized in terms of the  activities as evolving in time.

As to the first issue, consider the joint probability $P(k, e)$ where $k$ can now be considered as a discrete RV indexing the streams and $e$ is a discrete RV indexing the given activity set (\emph{argue within two feet}, etc.). Such joint distribution can be empirically estimated as   $N(k = i, e = j) / \sum_{k} \sum_{e}N(k , e )$, where $N(k = i, e = j)$ denotes the number of frames of stream $i$ that displays activity $j$. In Fig.~\ref{fig:events_distr}, $P(k, e)$ is rendered as a 2D Hinton diagram.  The joint distribution is suitable to provide two essential pieces of information.

On the one hand, the marginalization of $P(k, e)$  over $e$, i.e.,  $P(k)= \sum_e P(k, e)$, offers an insight into the relevance of each stream - in terms of the marginal likelihood $P(k)$ - to the job of collating informative  stream subsets (cfr.  Fig.~\ref{fig:events_distr}). Intuitively, we expect  the corresponding marginal computed  after stream processing, say $\widetilde{P}(k)$,  to be a sparse summarisation of $P(k)$.  Yet,   it should account for the original representational relevance of the streams  (cfr.  Fig.~\ref{fig:event-distr_out}). This can be further understood by displaying the distribution of activities across cameras and time as in Fig.~\ref{fig:gt} (a darker color indicates several activities co-occurring at the same time). By comparing with the Hinton diagram of $P(k)$ in   Fig.~\ref{fig:events_distr}, it is readily seen that some cameras capture a large amount of  activities - for instance $cam 16$ and $cam 31$ -, whilst other cameras, e.g., $cam 20$ and $cam 27$, feature  few activities. At the same time, the information displayed by subsets of one stream can be considered as redundant with respect to subsets of another stream (e.g., $cam 37$ with respect to $cam 31$, Fig.~\ref{fig:gt}).

On the other hand, by marginalizing over $k$,  the  distribution of the activities in the data set is recovered, i.e. $P(e)= \sum_k P(k, e)$. It can be noted that (cfr.  Fig.~\ref{fig:events_distr}) such distribution is not uniform: some activities are  under-represented compared to other activities. This class imbalance problem entails two issues.  First, any kind of processing performed to select subsets of the video streams for  collating the most relevant data and information of interest should preserve the shape of such distribution, i.e. $\widetilde{P}(e) \approx P(e)$, where $\widetilde{P}(e)$ is the marginal distribution after processing.  Second, non uniformity should be accounted for when defining quantitative evaluation measures~\cite{he2013imbalanced}. Indeed, a suitable metric should reveal the true behavior of the method over minority and majority activities: the assessments of over-represented and under-represented activities should contribute equally to the assessment of the whole method. To cope with such a problem we jointly use two assessment metrics:  the  \emph{standard accuracy} and the \emph{macro average accuracy}~\cite{he2013imbalanced}.

Denote: $NP_{e}$ the number of positives, i.e.,  the number of times  the activity $e$ occurs in the entire recorded scene, independently of the camera;  $TP_{e}$ the number of \emph{true positives} for  activity $e$, i.e.,  the number of frames of the output video sequence that contain  activity $e$. Given $NP_{e}$ and $TP_{e}$ for each activity,  the following  can be defined.
\begin{itemize}
\item \emph{Standard Accuracy} 
$A = \frac{\sum_{e=1}^{E} TP_{e}}{\sum_{e=1}^{E} NP_{e}}$.
Note that this is a global measure that does not take into account the accuracy achieved on a single activity. From now on we will refer to this metric simply as \emph{accuracy}.
\item \emph{Macro Average Accuracy} 
$avg(A) = \frac{1}{E}\sum_{e=1}^{E} A_{e} = \frac{1}{E}\sum_{e=1}^{E} \frac{TP_{e}}{NP_{e}}$.
This  is the arithmetic average of the partial accuracy $A_{e}$ of each activity. It  
 allows each partial accuracy to contribute equally to the method assessment. We will refer to this metric simply as \emph{average accuracy}.
\end{itemize}

\subsection{Parameters and experiments setup}

We used $1000$ frames to setup strategy parameters: 
\begin{itemize}
\item \emph{Bayesian}: the initial hyper-parameters $\nu^{(k)}_0, \Delta^{(k)}_0$;
\item \emph{Random}: the parameter $b_w$ of the probability distribution;
\item \emph{Deterministic}: the within-stream time parameter $\Delta_w$ that modulates camera switches;
\item \emph{Charnov}: the slope of the gain function $\delta$.
\end{itemize}
The remaining $7000$ frames have been used for testing giving-up strategies against the  different visual representations previously introduced.

Note that a further constraint  is to be taken into account for a fair performance assessment. 
 In foraging terms,  it is the number of times the forager  chooses to explore a new patch; namely the number of camera switches. 

While in general, the estimated parameters are those that maximize performance of a method, here parameters have been selected so to maximize  the accuracy (or average accuracy) while keeping the number of camera switches below a given boundary condition.  A measure of the accuracy of the system, should not be given as an absolute value, but the selection of   subsets of the video streams should be performed  to collate a ``meaningful''  summary in which the switching frequency is bounded. To thoroughly address this surmise, the accuracy behavior  as function of the number of camera switches has been studied. The overall result can be summarised as: first, all representation schemes, apart from S, combined with the Bayesian giving-up strategy, achieve their best performance at a small number of camera switches; second, all  giving-up strategies combined with the visual information method M achieve their best performance at a small number of camera switches  (cfr., Fig. 14a and 14b of Supplementary Material). 
That being the experimental evidence, a reasonable upper bound can  be  determined either by taking into account  the intrinsic limitations of the human visual system  and/or, semantically, the characteristics of  time activity distribution in the dataset. 
As to the first issue, consider that human subjects  looking at videos explore the scene through saccadic eye-movements with maximal saccade duration of approximately $160$ ms and  $340$ ms average post-saccadic fixational time (when cognitive processing takes place)\cite{dorr2010variability}. Post-saccadic exploration can be even longer in case of pursuit (depending on task).  Thus, a reasonable  time to be granted for  visual foraging  is approximately  one second (e.g, one/two saccades followed by pursuit, or two/tree control saccades with brief fixations). The length of the test scene is about $7000$ frames,  $30$ fps frame rate, thus a conservative upper bound for the number of  switches is about $240$. This is somehow consistent  with empirical analysis of accuracy over switch number, where above $300$ camera switches strategies become comparable to the random strategy, in some cases worse. Under the circumstances, we slightly relax the upper bound to $280$.

As regards activity duration, note that the average length of each activity occurring in the scene across cameras is about $500$ frames. Ideally,   a camera switch should take place after having observed  a full activity. Thus,  given a  stream length of $7000$ frames, an upper bound estimate  for the number of camera switches is about $14$.
Since each boundary condition might determine a different set of parameters, distinct learning  and testing phases have been performed for each boundary condition, that is  \#cam\_switch $<280$ and  \#cam\_switch $<14$.

\subsection{Results}

Table~\ref{table:results1} and  Table~\ref{table:results2} report  quantitative assessment of results achieved by the different foraging strategies dependent on the available  visual representations. 
Table~\ref{table:results1} has been obtained  by considering  the upper bound \#cam\_switch $< 280$; Table~\ref{table:results2}  relates to the condition \#cam\_switch $<14$.

Beyond the fact that the M/Bayesian scheme, at the core of the proposed model, overcomes other schemes both in terms of accuracy and average accuracy, some interesting results are worth a comment. 

First, the proposed Bayesian giving-up strategy performs better than other strategies, in terms of both standard and average accuracy,  independent of the visual representation adopted. At the same time, it is not affected by the chosen upper bound on the number of camera switches, whilst for other strategies the ``semantic'' upper bound (\#cam\_switch $<14$) pairs with a slight decrease in performance. Both results confirm that for  the task of monitoring multiple cameras, the stream switching strategy is a  crucial issue, which might drastically affect the overall performance.

Second,  the behavior of the monitoring system  is not agnostic about the visual representation adopted. Results reported in both conditions, give quantitative evidence of the higher representation capability of    M and M+  as opposed to S and S+CD+FB, which could be qualitatively appreciated by simple visual inspection of the $\mathcal{C}^{(k)}(t)$ behavior (graphs shown in Supplementary Materials). This holds independently of the strategy followed. Such effect is more clear by considering results obtained via the Random strategy. Recall that the latter selects a new stream after a within-stream time interval $\Delta_w$, which is randomly sampled. However, the camera content is not selected by chance and the quality of the visual complexity index $\mathcal{C}^{(k)}(t)$ plays a fundamental role for determining stream selection.

Third, the best performance achieved by M with respect to M+, shows that  the ideal observer who behaves in an uncertain environment as an ``omniscient forager'' (switching is decided by surmising full knowledge of objects in streams \cite{stephens1986foraging}) is likely to perform less efficiently than a ``prudent'' one. This, together with the remarkable difference between  results achieved via Charnov and Bayesian strategies,  confirms the inadequacy of a pure Charnov strategy in  complex experimental setting \cite{wolfe2013time}. Similar considerations could be formulated on the other results reported as Supplementary Material.

\begin{figure}[tb]
  \centering
    \includegraphics[width=0.40\textwidth]{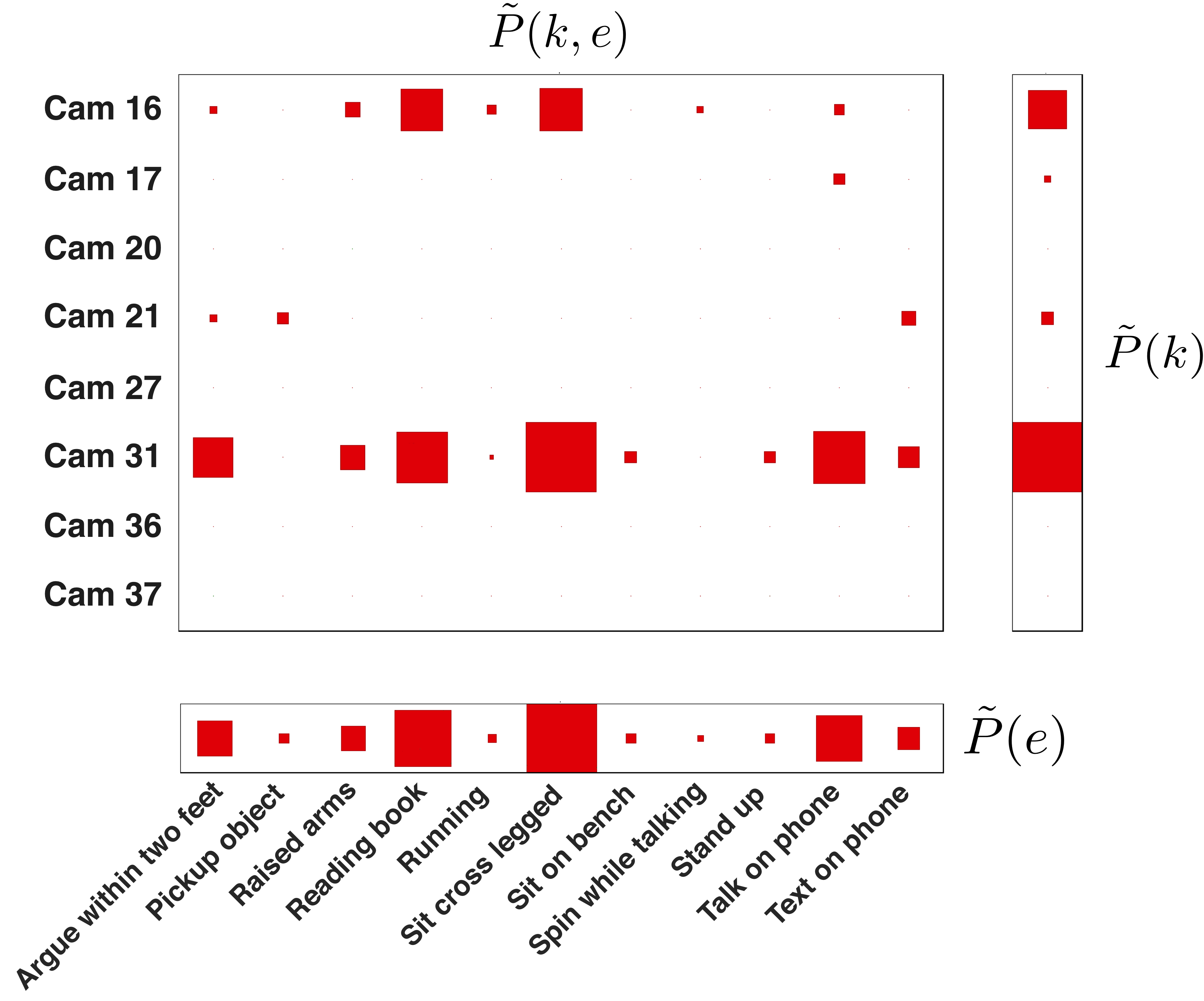}
  \caption{Distribution  of activities in terms of joint and marginal probability distributions $\tilde{P}(k,e)$, $\tilde{P}(k)$ and $\tilde{P}(e)$, respectively,  obtained after processing   (M representation and Bayesian foraging strategy  with optimal \#cam\_switch $<$ 14). To be compared with  Fig.~\ref{fig:events_distr}.}
  \label{fig:event-distr_out}
\end{figure}

At a glance, the overall behavior of the M/Bayesian scheme  can be appreciated in Fig.~\ref{fig:event-distr_out}, which is  equivalent to the representation provided in Fig.~\ref{fig:events_distr}, but computed on the output collated stream.
\begin{table}[tb]
\tiny
\centering
\begin{tabular}{lccccc}
& \multicolumn{4}{c}{\bf{Foraging strategy}} \\
\cline{2-6}
\bf{Visual Information}	&	\bf{Measure}	&	Random	&	Deterministic	&	Charnov	&	Bayesian \\
\hline
M	&	accuracy	&	60.11	&	68.52	&	61.83		&	\bf{82.53}		\\
	&	avg\_accuracy	&	45.77	&	52.09	&	48.20				&	\bf{71.12}		\\
\hline
M+	&	accuracy	&	59.35	&	67.33	&	53.33			&	77.06	\\
	&	avg\_accuracy	&	43.95	&	58.32	&	45.55			&	57.56		\\
\hline
S	&	accuracy	&	16.24	&	19.89	&	24.40		&	79.29		\\
	&	avg\_accuracy	&	14.33	&	19.45	&	14.98	&	70.79		\\
\hline
S+CD+FB	&	accuracy	&	21.17	&	24.64	&	35.61		&	82.08		\\
	&	avg\_accuracy	&	19.36	&	24.56	&	21.95		&	68.47		\\
\hline
\end{tabular}
\medskip 
\caption{Accuracy and average accuracy achieved by the baseline and proposed foraging strategies combined with several visual  processing methods. Results have been obtained for  \#cam\_switch $<$ 280. Best performance are reported in bold.}
 \label{table:results1}
 \end{table} 
\begin{table}[!]
\tiny
\centering
\begin{tabular}{lccccc}
& \multicolumn{4}{c}{\bf{Foraging strategy}} \\
\cline{2-6}
\bf{Visual Information}	&	\bf{Measure}	&	Random	&	Deterministic	&	Charnov	&	Bayesian \\
\hline
M	&	accuracy	&	50.04	&	54.18	&	59.83&	\bf{88.74}		\\
	&	avg\_accuracy	&	33.08	&	29.60	&	34.62				&	\bf{76.99}		\\
\hline
M+	&	accuracy	&	50.99	&	56.55	&	55.09			&	80.72	\\
	&	avg\_accuracy	&	34.06	&	35.49	&	29.50			&	72.98		\\
\hline
S &	accuracy	&	16.24	&	20.58	&	41.42		&	59.15		\\
	&	avg\_accuracy	&	14.33	&	14.98	&	22.85	&	50.54		\\
\hline
S+CD+FB	&	accuracy	&	19.95	&	20.66	&	26.09	&		74.98		\\
	&	avg\_accuracy	&	16.23	&	14.91	&	19.48		&	51.05		\\
\hline
\end{tabular}
\medskip 
\caption{Accuracy and average accuracy for  \#cam\_switch $<$ 14.}
\label{table:results2}
 \end{table} 
The joint  distribution $\widetilde{P}(k, e)$ and the marginal $\widetilde{P}(k)$  appear, as expected, to be a sparse summarisation of the initial dataset distributions $P(k, e)$ and $P(k)$   shown in Fig.~\ref{fig:events_distr}. In particular, apart from \emph{cam37}, the Bayesian strategy selects the most informative cameras, e.g., \emph{cam16}, \emph{cam31}, while discarding the less informative ones, e.g., \emph{cam20}, \emph{cam27}. Most important, the model provides an output marginal distribution of  activities $\widetilde{P}(e)$ that is very close to the initial distribution $P(e)$. This result shows  that by exploiting the proposed approach, most relevant activities have been captured.

\begin{figure}[tb]
  \centering
    \includegraphics[width=0.34\textwidth]{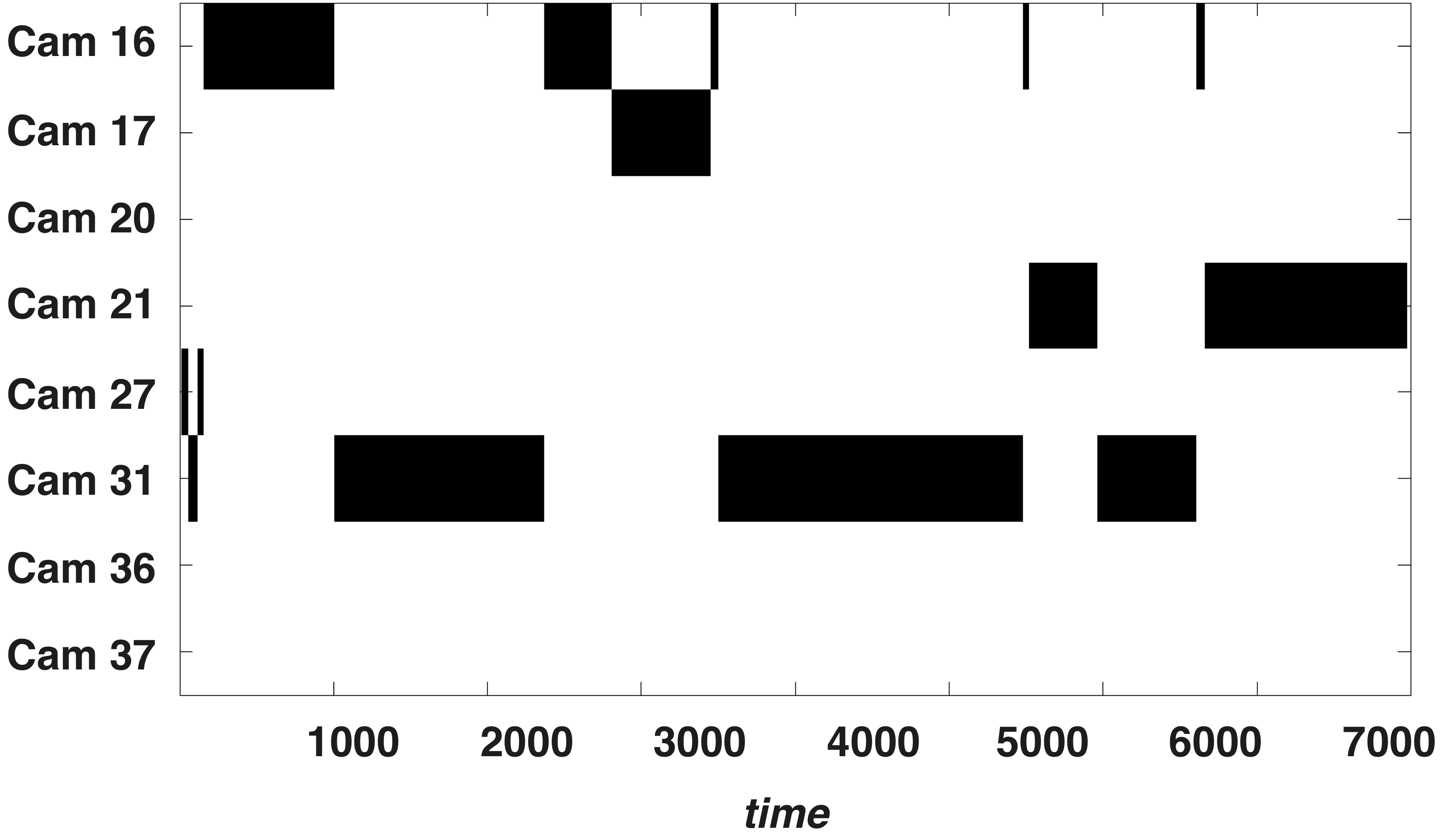}
  \caption{Output timeline via the M/Bayesian scheme. It shows the sequence of the video stream subsets collated by the proposed model  in terms of camera switches. At each time, the corresponding camera content is represented as a black rectangle and the output  stream includes the content of one camera.}
  \label{fig:out_timeline}
\end{figure}

\begin{figure}[tb]
  \centering
  \scriptsize
   \setlength{\tabcolsep}{2.1pt}
  \begin{tabular}{cccc}
  \includegraphics[width=0.1\textwidth]{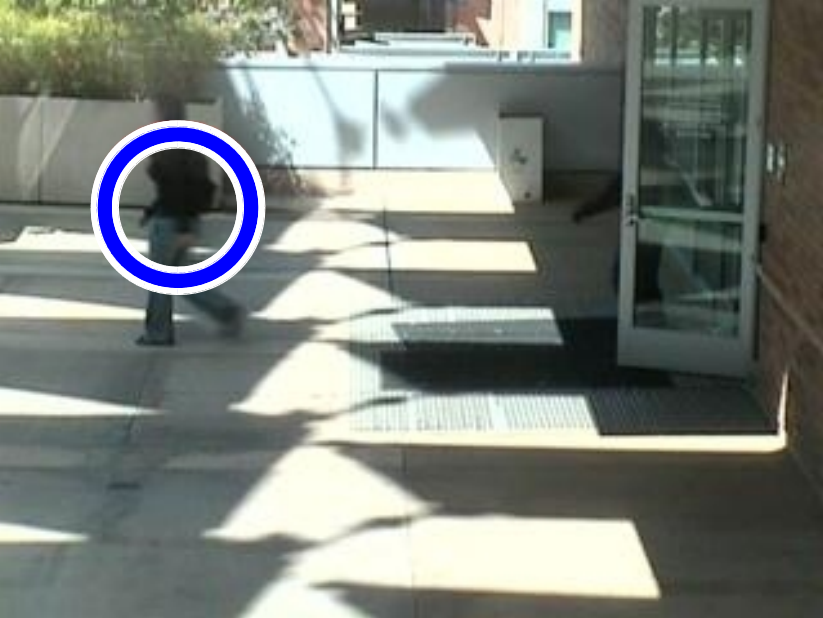} &
  \includegraphics[width=0.1\textwidth]{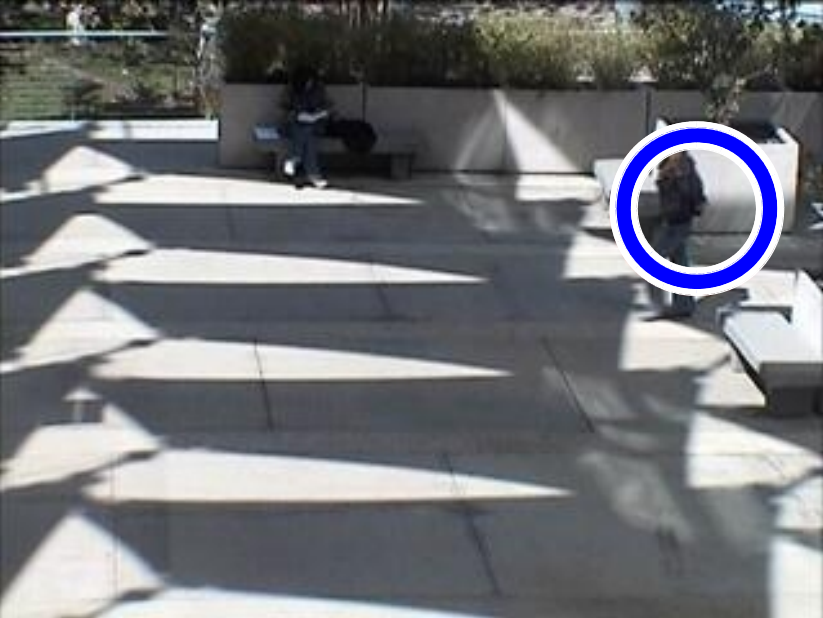} &
  \includegraphics[width=0.1\textwidth]{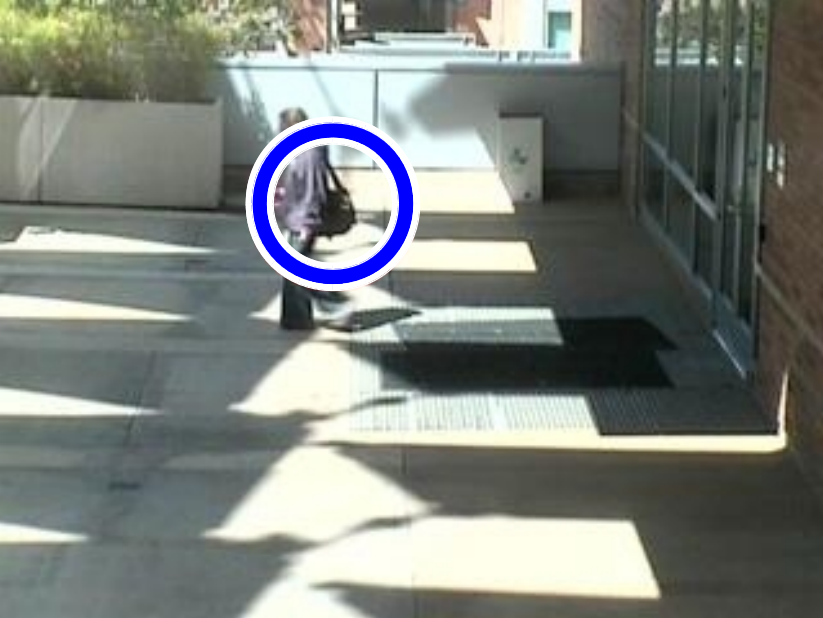} &
   \includegraphics[width=0.1\textwidth]{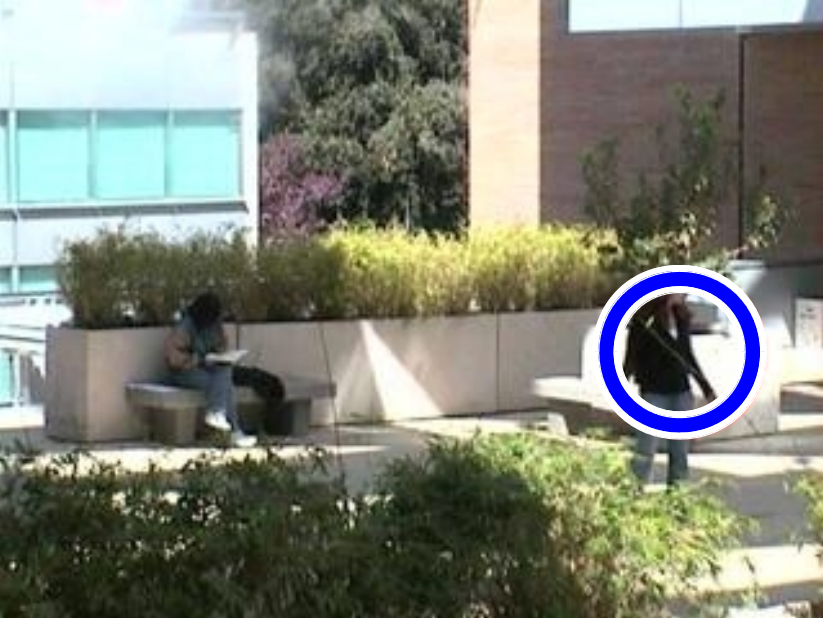}\\
  (t=1,cam=27)&(t=58,cam=31)&(t=119,cam=27)&(t=157,cam=16)\\
  \includegraphics[width=0.1\textwidth]{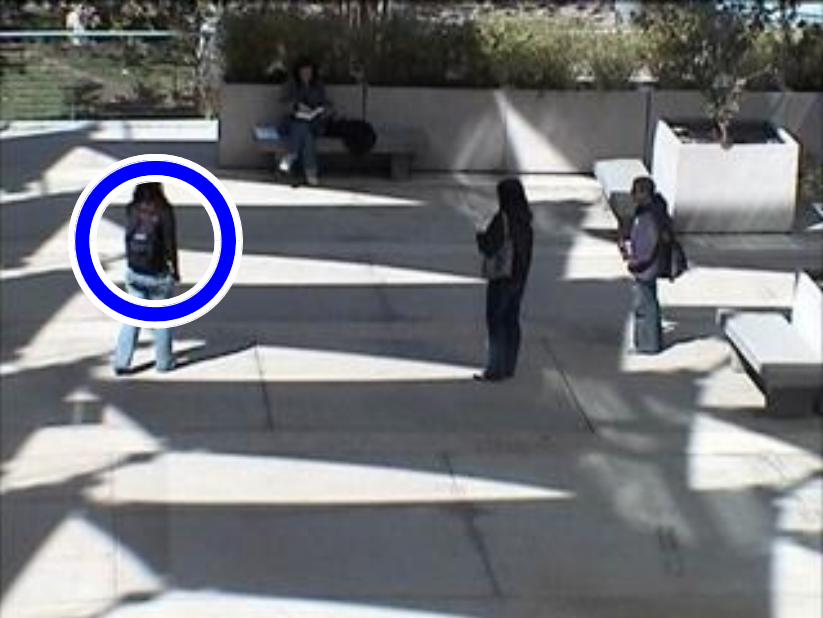} &
     \includegraphics[width=0.1\textwidth]{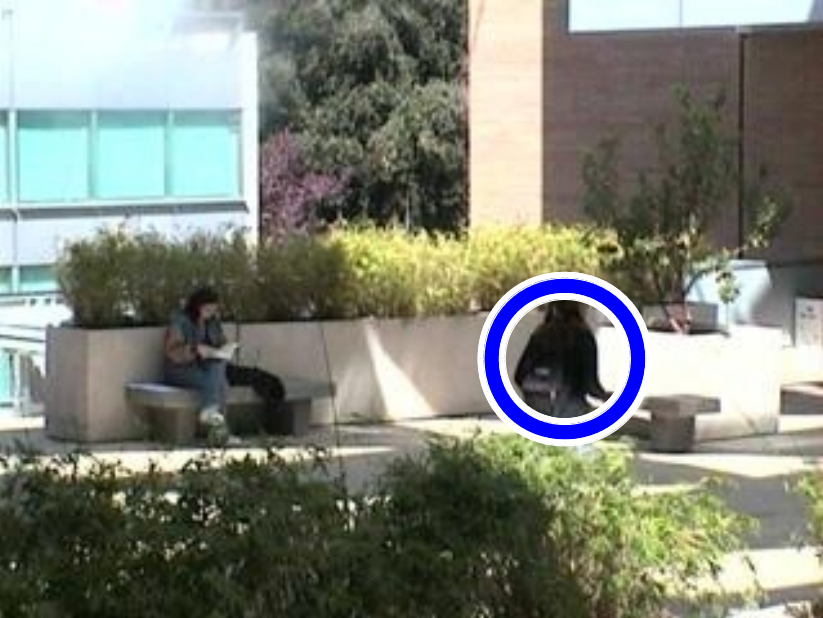} &
  \includegraphics[width=0.1\textwidth]{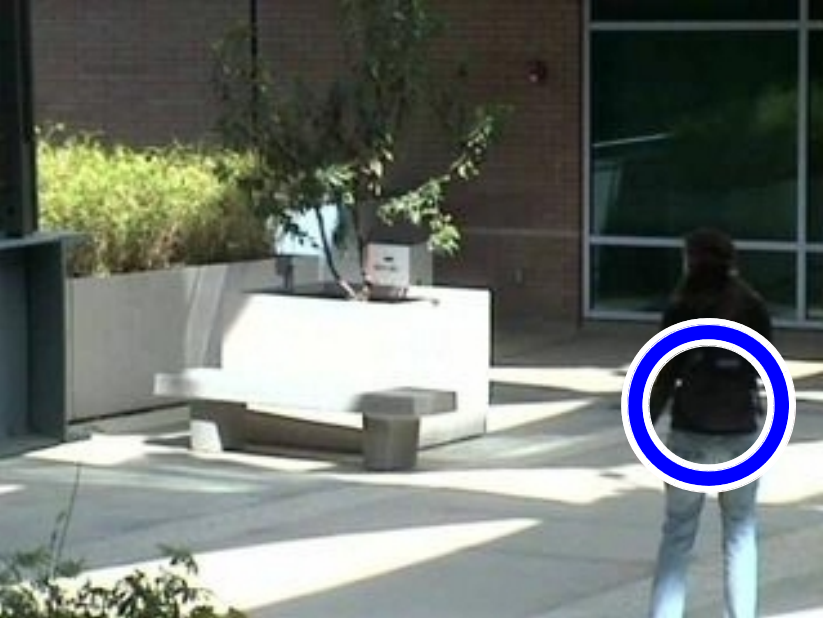} &
  \includegraphics[width=0.1\textwidth]{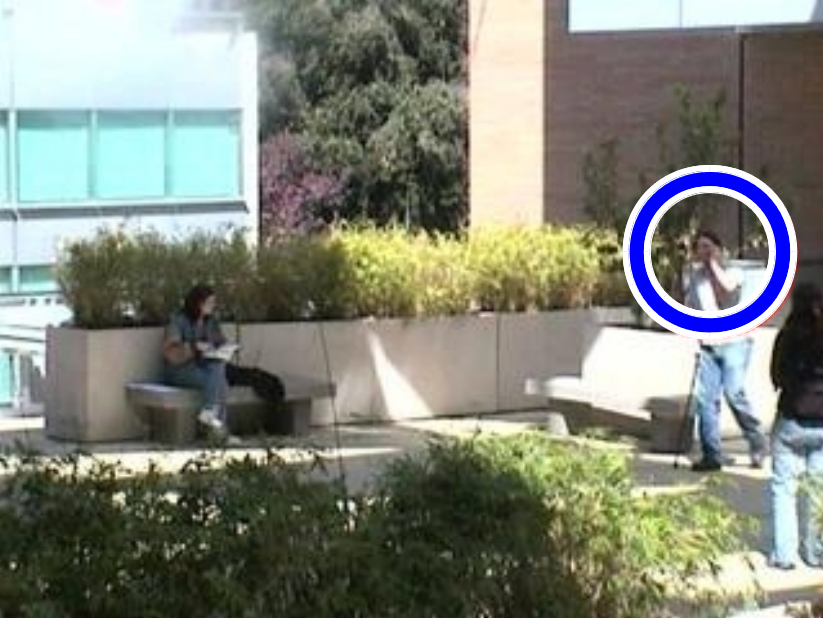} \\ 
  (t=1007,cam=31)&(t=2369,cam=16)&(t=2808,cam=17)&(t=3452,cam=16)\\     
  \includegraphics[width=0.1\textwidth]{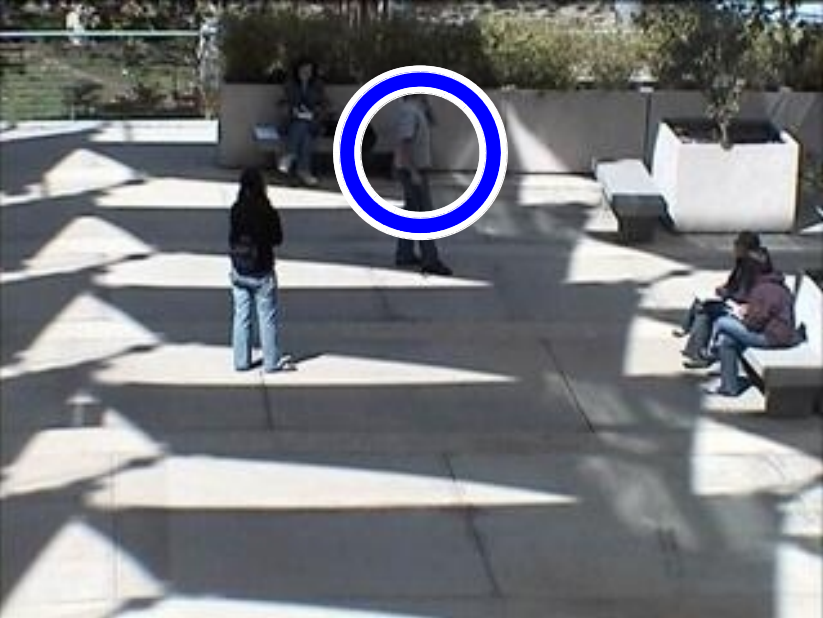} &
    \includegraphics[width=0.1\textwidth]{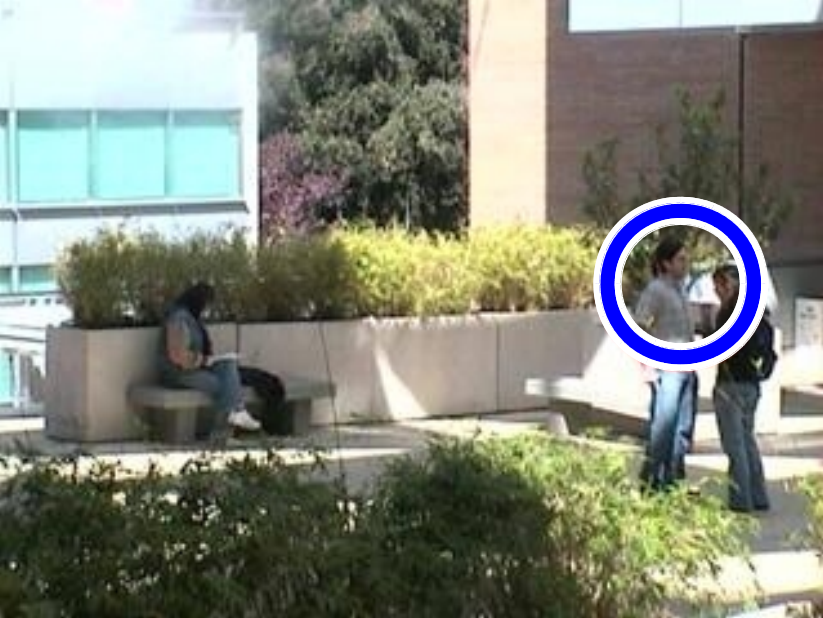} &
   \includegraphics[width=0.1\textwidth]{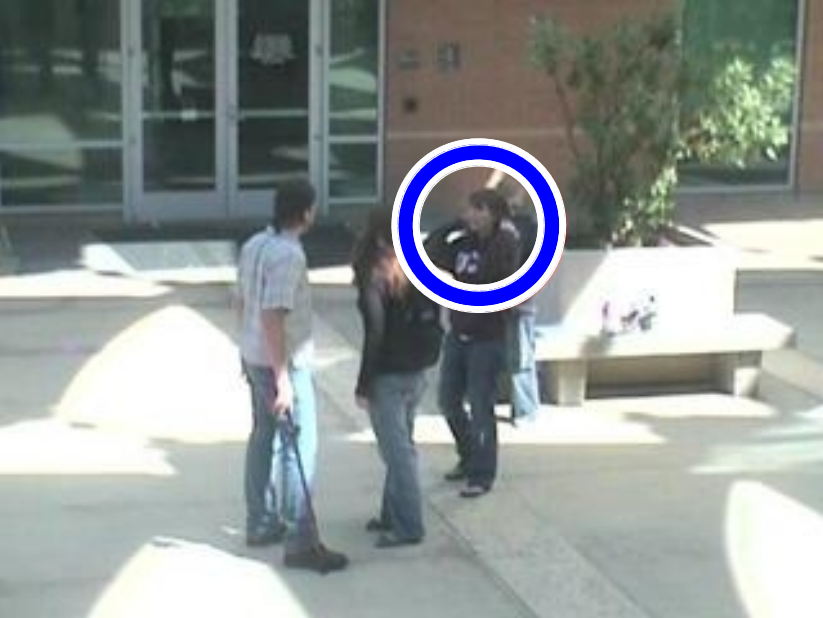} &
  \includegraphics[width=0.1\textwidth]{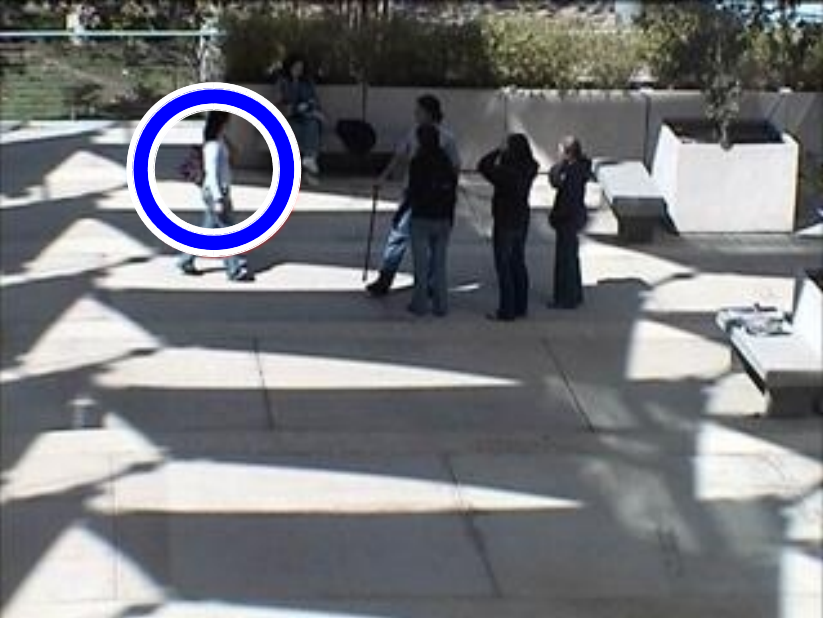} \\
(t=3499,cam=31)&(t=5480,cam=16)&(t=5517,cam=21)&(t=5960,cam=31)\\
    \includegraphics[width=0.1\textwidth]{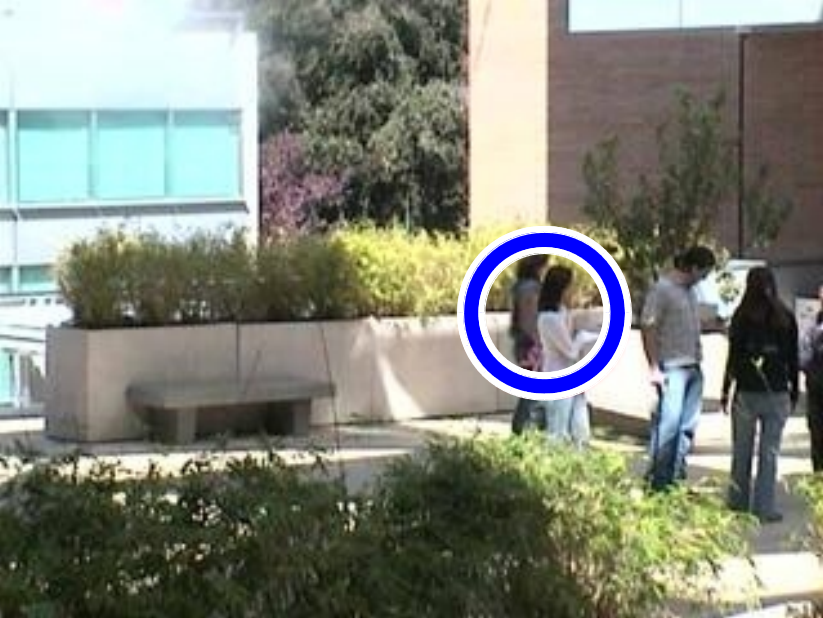} &
  \includegraphics[width=0.1\textwidth]{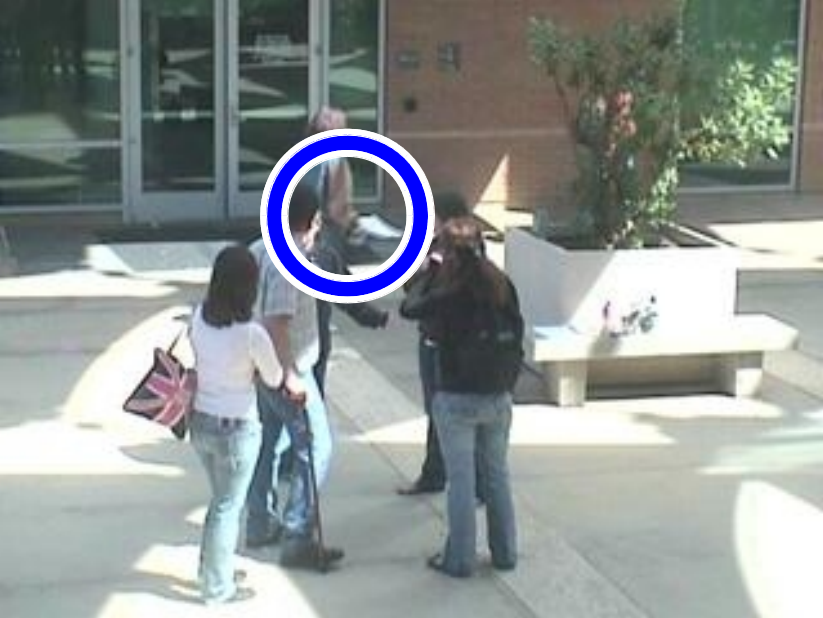}  & &\\
  (t=6604,cam=16)& (t=6661,cam=21)& &\\   
   \end{tabular}
  \caption{A typical output of the M/Bayesian scheme, which recaps the foraging activity: the subset of frames extracted   captures the most important data for the surveillance analysis task. Here, the output sequence is summarised  through the camera switches performed by the optimal Bayesian observer. From the first at t=1 (top-left) to the last at t=6661 (bottom-right). In each camera frame the actual FoA is displayed as a white/blue circle.}
  \label{fig:foa_output}
\end{figure}

Eventually, the concrete output of the M/Bayesian scheme's  can be summarised in terms of the   timeline  visualized in Fig.\ref{fig:out_timeline}. This represents the final ``storyboard'', i.e., the video stream subsets collated by the proposed model  in terms of camera switches. Here the output  stream contains only one camera content at each time,  and it can be considered as a ``binarized'' version of Fig.~\ref{fig:gt}. From such representation,  the example  presented  in Fig.~\ref{fig:foa_output} can be recovered.  The sequence of most important frames can be thought of as a new composed video obtained by  sequentially switching from one camera  to another.   As it can be observed (Fig.~\ref{fig:foa_output}), each camera switch has been triggered by human activities, e.g. \emph{walking} at t=1, \emph{spin while talking} at t=2808, \emph{raised arms} at at t=6661, etc.

\section{Final remarks and Conclusion}
\label{sec:final}

We have presented a  unifying theoretical framework for  selecting  subsets of multiple video streams for  collating the most relevant data and information of interest related to a given task. The  framework  formulates  attentive monitoring  as the behavior of a forager  that, moment to moment, focuses  his attention on the most informative stream/camera, detects interesting objects for the task at hand, switches from the current stream to a more informative one.
Experimental results achieved on the UCR Videoweb Activities Dataset, have been presented to assess the performance of the proposed technique.
To the best of our knowledge the model proposed here is novel for the multi-camera surveillance research field. 

The  approach  could be either  straightforwardly exploited for i) reducing the manual operator fatigue for multiple monitor situation or ii) as a preliminary step for intelligent  surveillance relying on the analysis of  actions, activities and behaviors.
There are however some current limitations in the  model that should be  addressed for on-field application.
 
As to the first scenario, the actual visual foraging  of a human operator should be taken into account  for learning model parameters,   which should entail  two steps. First, mobile eye-tracking of operator's gaze behavior can be performed in the experimental setting  of a typical  control center with the human  engaged in inspecting  a number of camera monitors on the wall. There are few experiments of this sort in the psychological literature (e.g. \cite{wolfe2013time}) but limited to  simple target visual search. In the present work, eye-tracking data  from human subjects have been used, but  limited to the inference of parameters of distributions related to  oculomotor biases;  namely, the prior for sampling gaze shifts within the stream ~\cite{TatlerBallard2011eye,BocFerSMCB2013}.  Second, the components of the model should be implemented in order to allow full learning. For what regards the visual component (Fig.\ref{fig:sense}), in a time-varying perspective, it can be conceived as a time slice of a Dynamic Bayesian Network; then, distribution parameters can be learned with a variety of methods available \cite{koller2009probabilistic}. Foraging parameters of the executive component can be inferred using optimization methods that have been proposed  for dealing with actual forager behaviors in a variety of patch/prey conditions~\cite{mcnamara2006bayes}.

For what concerns  high-level intelligent analysis,  current implementation of the model focuses on local object analysis and does not consider different levels of semantic information captured by cameras with different scales and angles. Attentive modeling of actions, activities and behaviors is a hot field in computer vision   and results obtained up to now could be integrated within our framework with moderate effort. As it has been shown in the experimental analyses, the model offers a probabilistic framework in which it is easy to accommodate  a variety of available state-of-the-art attention-based algorithms \cite{BorItti2012}. Further, note that the Bayesian strategy (Eq. \ref{eq:opt}) basically relies on the configurational complexity $\mathcal{C}^{(k)}(t)$, which, in turn, is based on spatial entropy. The latter is a mesoscopic quantity that summarises and can be derived from a variety of ''atomic " visual measures (e.g, see \cite{Regazzoni_AVSS2013}). However, from a strict engineering perspective much depends on the specific field of application that is to be addressed.  

In the specific case of multi-stream summarisation, for instance, the method can be used as such, similarly to Kankanhalli \emph{et al.}~\cite{kankanhalli2006expB}. Alternatively, it is suitable to provide a principled base to approaches such as those performing correspondence-free multi-camera activity analysis~\cite{Wang2013}. 

An interesting issue is the applicability of the approach to the case of online multi-camera  systems. This case compels to take into account the architectural complexities of the network. The latter  can be factored  in terms of  distribution, mobility and degree of motion of the sensors \cite{CucchiaraMobile2010}. As to the distribution issue, the pre-attentive control loop is suitable to be considered for a straightforward fully decentralized implementation, while the attentive loop could be designed  at different levels of distribution. Interestingly enough, the  ecological Bayesian handoff mechanism is suitable to embed resource-aware conditions, e.g., energy consumption, that are also considered in actual animal foraging.
For what concerns the degree of motion of the sensors, clearly, the visual attention  rationale that is behind our model calls for considering smart camera networks embedding PTZ cameras that are able to dynamically modify their FOV. In this case, for what concerns single camera activities,  techniques developed in the active vision field  are apt to be embedded in within-frame analysis either at the pre-attentive or the attentive stage \cite{MichelForesti2010}.  However,   at some point multi-camera activity analysis requires fusing information from multiple camera views. The data fusion problem has not been explicitly considered in this paper. Yet observational data may be combined, or fused, at a variety of levels \cite{hall1997introduction}, again depending on the architecture devised for a specific application.  

In ongoing  research we are  considering multimodal data-fusion at the sensor level for  audio/video integration in  ambient intelligence. To such end the perceptual component can be straightforwardly extended to cope with other sources of information. Indeed,  crossmodal integration can apparently arise before attentional selection is completed \cite{driver1998crossmodal}, which
can be accounted for by exploiting the priority map  for representing cross modal integration at this level. Addressing fusion at higher levels calls for software architecture abstractions to allow components to interact even if they rely on different spatial models. In this perspective, we are adapting a framework of  space-based communication, to serve as an architectural  support for foraging in augmented ecologies \cite{tisato2012grounding}. 

\appendices
\section{Proof of Proposition \ref{prop:strategy}}
\label{app:opt}
For an optimal Bayesian   observer the decision to leave  the current stream is based  on the posterior probability that a reward can be gained within the stream (complexity), given that no reward has been gained by time $t$;  via Bayes' rule:
\begin{equation}
P(\mathcal{C}^{(k)}(t) \mid \neg R^{(k)}(t)) = \frac{P(\neg R^{(k)}(t) \mid \mathcal{C}^{(k)}(t))P(\mathcal{C}^{(k)}(t))}{P(\neg R^{(k)}(t))}
\label{eq:post}
\end{equation}
\noindent where $P(\neg R^{(k)}(t)) = 1 - P(R^{(k)}(t))$, $P(R^{(k)}(t))$ denoting the marginal likelihood of being rewarded. Using the detection function, Eq.\ref{eq:gain}, the likelihood of not gaining  reward is 
\begin{equation}
P(\neg R^{(k)}(t) \mid \mathcal{C}^{(k)}(t))  =  \exp(-\lambda t).
\label{eq:ngain}
\end{equation} 
Since, by definition,  reward can  be actually gained only within the currently visited stream,
\begin{equation}
\begin{split}
&P(R^{(k)}(t))  = \sum_{ \mathcal{C}(t) \in \{  \mathcal{C}^{(k)} \}_{k=1}^{K} } P(\mathcal{C}(t))P(R^{(k)}(t) \mid \mathcal{C}(t)) =\\
& P(\mathcal{C}^{(k)}(t)) P(R^{(k)}(t) \mid \mathcal{C}^{(k)}(t)).
\end{split}
\label{eq:marg}
\end{equation}
Taking into account that $P(\neg R^{(k)}(t)) = 1 - P(R^{(k)}(t)) $,  the definition of the detection function, Eq. \ref{eq:gain} and  Eq. \ref{eq:marg}
\begin{equation}
P(\neg R^{(k)}(t))  =  1 - P(\mathcal{C}^{(k)}(t)) (1- \exp(-\lambda t)).
\label{eq:nmarg}
\end{equation}
By the total law of probability, $1 - P(\mathcal{C}^{(k)}(t)) = \sum_{i \neq k}P(\mathcal{C}^{(i)}(t))$, thus previous equation can be written as
\begin{equation}
P(\neg R^{(k)}(t))  =  \sum_{i \neq k}P(\mathcal{C}^{(i)}(t)) - P(\mathcal{C}^{(k)}(t)) \exp(-\lambda t).
\label{eq:nmarg}
\end{equation}
Plugging into the posterior (Eq. \ref{eq:post}) and rearranging
\begin{equation}
\begin{split}
&P(\mathcal{C}^{(k)}(t) \mid \neg R^{(k)}(t)) =\\
& \frac{P(\mathcal{C}^{(k)}(t))}{\exp(\lambda t)\sum_{i \neq k}P(\mathcal{C}^{(i)}(t)) - P(\mathcal{C}^{(k)}(t)) }.
\end{split}
\label{eq:post2}
\end{equation}
Optimal behavior consist in switching when the posterior is equal for  all streams,  thus
\begin{equation}
\begin{split}
&\frac{1}{K} = 
\frac{P(\mathcal{C}^{(k)}(t))}{\exp(\lambda t)\sum_{i \neq k}P(\mathcal{C}^{(i)}(t)) - P(\mathcal{C}^{(k)}(t)) }\\
\end{split}
\label{eq:post2}
\end{equation}
\noindent which gives the condition:
\begin{equation}
K P(\mathcal{C}^{(k)}(t)) = \exp(\lambda t) \sum_{i \neq k}P(\mathcal{C}^{(i)}(t)) + P(\mathcal{C}^{(k)}(t)).
\label{eq:post3}
\end{equation}
Rearranging terms,  
using the prior probability $P(\mathcal{C}^{(k)}(t))$, Eq.  \ref{eq:P_C}, and inserting in Eq. \ref{eq:post3}, then the optimal condition for stream leaving, boils down to
\begin{equation}
 \mathcal{C}^{(k)}(t) \exp(-\lambda t) =  \frac{1}{K-1}\sum_{i \neq k}\mathcal{C}^{(i)}(t),
\label{eq:switch2}
\end{equation}
\noindent which proofs Eq. \ref{eq:opt}.

\section*{Acknowledgments}
The authors are grateful to the Referees and the Associate Editor, for  enlightening remarks  that have greatly improved the quality  of an earlier version of this paper, and to Prof. Raimondo Schettini for having inspired the present work.

\bibliographystyle{IEEEtran}

\bibliography{./IEEEabrv,./attentive_abbr,./levyeye_abbr}

\begin{thebibliography}{10}
\providecommand{\url}[1]{#1}
\csname url@samestyle\endcsname
\providecommand{\newblock}{\relax}
\providecommand{\bibinfo}[2]{#2}
\providecommand{\BIBentrySTDinterwordspacing}{\spaceskip=0pt\relax}
\providecommand{\BIBentryALTinterwordstretchfactor}{4}
\providecommand{\BIBentryALTinterwordspacing}{\spaceskip=\fontdimen2\font plus
\BIBentryALTinterwordstretchfactor\fontdimen3\font minus
  \fontdimen4\font\relax}
\providecommand{\BIBforeignlanguage}[2]{{%
\expandafter\ifx\csname l@#1\endcsname\relax
\typeout{** WARNING: IEEEtran.bst: No hyphenation pattern has been}%
\typeout{** loaded for the language `#1'. Using the pattern for}%
\typeout{** the default language instead.}%
\else
\language=\csname l@#1\endcsname
\fi
#2}}
\providecommand{\BIBdecl}{\relax}
\BIBdecl

\bibitem{stephens1986foraging}
D.~W. Stephens, \emph{Foraging theory}.\hskip 1em plus 0.5em minus 0.4em\relax
  Princeton University Press, 1986.

\bibitem{xiang2006beyond}
T.~Xiang and S.~Gong, ``Beyond tracking: Modelling activity and understanding
  behaviour,'' \emph{Int. J. Comput. Vis.}, vol.~67, no.~1, pp. 21--51, 2006.

\bibitem{hills2006animal}
T.~T. Hills, ``Animal foraging and the evolution of goal-directed cognition,''
  \emph{Cognitive Science}, vol.~30, no.~1, pp. 3--41, 2006.

\bibitem{wolfe2013time}
J.~M. Wolfe, ``When is it time to move to the next raspberry bush? foraging
  rules in human visual search,'' \emph{J. Vis.}, vol.~13, no.~3, p.~10, 2013.

\bibitem{charnov1976optimal}
E.~L. Charnov, ``Optimal foraging, the marginal value theorem,''
  \emph{Theoretical population biology}, vol.~9, no.~2, pp. 129--136, 1976.

\bibitem{schutz2011eye}
A.~Sch{\"u}tz, D.~Braun, and K.~Gegenfurtner, ``Eye movements and perception: A
  selective review,'' \emph{Journal of Vision}, vol.~11, no.~5, 2011.

\bibitem{fuster2004upper}
J.~M. Fuster, ``Upper processing stages of the perception--action cycle,''
  \emph{Trends in cognitive sciences}, vol.~8, no.~4, pp. 143--145, 2004.

\bibitem{fuster2009cortex}
------, ``Cortex and memory: emergence of a new paradigm,'' \emph{J. of
  Cognitive Neuroscience}, vol.~21, no.~11, pp. 2047--2072, 2009.

\bibitem{haykin2014cognitive}
S.~Haykin and J.~M. Fuster, ``On cognitive dynamic systems: Cognitive
  neuroscience and engineering learning from each other,'' \emph{Proc. {IEEE}},
  vol. 102, no.~4, pp. 608--628, 2014.

\bibitem{chiappino2014bio}
S.~Chiappino, P.~Morerio, L.~Marcenaro, and C.~S. Regazzoni, ``Bio-inspired
  relevant interaction modelling in cognitive crowd management,'' \emph{J. of
  Ambient Intell. and Humanized Computing}, pp. 1--22, 2014.

\bibitem{chiappino2014event}
S.~Chiappino, L.~Marcenaro, P.~Morerio, and C.~Regazzoni, ``Event based
  switched dynamic bayesian networks for autonomous cognitive crowd
  monitoring,'' in \emph{Wide Area Surveillance}.\hskip 1em plus 0.5em minus
  0.4em\relax Springer, 2014, pp. 93--122.

\bibitem{dore2010interaction}
A.~Dore, A.~F. Cattoni, and C.~S. Regazzoni, ``Interaction modeling and
  prediction in smart spaces: a bio-inspired approach based on autobiographical
  memory,'' \emph{{IEEE} Trans. Syst., Man, Cybern. {A}}, vol.~40, no.~6, pp.
  1191--1205, 2010.

\bibitem{Regazzoni_AVSS2013}
S.~Chiappino, L.~Marcenaro, and C.~Regazzoni, ``Selective attention automatic
  focus for cognitive crowd monitoring,'' in \emph{10th Int. Conf. Advanced
  Video Signal Based Surveillance}, Aug 2013, pp. 13--18.

\bibitem{Wang2013}
X.~Wang, ``Intelligent multi-camera video surveillance: A review,''
  \emph{Pattern Recognit. Lett.}, vol.~34, no.~1, pp. 3 -- 19, 2013.

\bibitem{leo2014multicamera}
C.~{De Leo} and B.~S. Manjunath, ``Multicamera video summarization and anomaly
  detection from activity motifs,'' \emph{ACM Trans. on Sensor Networks},
  vol.~10, no.~2, p.~27, 2014.

\bibitem{online_2015}
S.-H. Ou, C.-H. Lee, V.~Somayazulu, Y.-K. Chen, and S.-Y. Chien, ``On-line
  multi-view video summarization for wireless video sensor network,''
  \emph{{IEEE} J. Select. Topics Signal Processing.}, vol.~9, no.~1, pp.
  165--179, Feb 2015.

\bibitem{cong2012towards}
Y.~Cong, J.~Yuan, and J.~Luo, ``Towards scalable summarization of consumer
  videos via sparse dictionary selection,'' \emph{{IEEE} Trans. Multimedia},
  vol.~14, no.~1, pp. 66--75, 2012.

\bibitem{tron2011distributed}
R.~Tron and R.~Vidal, ``Distributed computer vision algorithms,'' \emph{{IEEE}
  Signal Processing Mag.}, vol.~28, no.~3, pp. 32--45, 2011.

\bibitem{MichelForesti2010}
C.~Micheloni, B.~Rinner, and G.~Foresti, ``Video analysis in pan-tilt-zoom
  camera networks,'' \emph{{IEEE} Signal Processing Mag.}, vol.~27, no.~5, pp.
  78--90, Sept 2010.

\bibitem{Roy-Chow_review2014}
A.~Kamal, C.~Ding, A.~Morye, J.~Farrell, and A.~Roy-Chowdhury, ``An overview of
  distributed tracking and control in camera networks,'' in \emph{Wide Area
  Surveillance}, ser. Augmented Vision and Reality, V.~K. Asari, Ed.\hskip 1em
  plus 0.5em minus 0.4em\relax Springer Berlin Heidelberg, 2014, vol.~6, pp.
  207--234.

\bibitem{qureshi2008smart}
F.~Qureshi and D.~Terzopoulos, ``Smart camera networks in virtual reality,''
  \emph{Proc. {IEEE}}, vol.~96, no.~10, pp. 1640--1656, 2008.

\bibitem{li2011utility}
Y.~Li and B.~Bhanu, ``Utility-based camera assignment in a video network: A
  game theoretic framework,'' \emph{{IEEE} Sensors J.}, vol.~11, no.~3, pp.
  676--687, 2011.

\bibitem{esterle2014socio}
L.~Esterle, P.~R. Lewis, X.~Yao, and B.~Rinner, ``Socio-economic vision graph
  generation and handover in distributed smart camera networks,'' \emph{ACM
  Trans. on Sensor Networks}, vol.~10, no.~2, p.~20, 2014.

\bibitem{Ballard}
D.~Ballard, ``Animate vision,'' \emph{Art. Intell.}, vol.~48, no.~1, pp.
  57--86, 1991.

\bibitem{aloimonos2013active}
Y.~Aloimonos, \emph{Active perception}.\hskip 1em plus 0.5em minus 0.4em\relax
  Psychology Press, 2013.

\bibitem{BorItti2012}
A.~Borji and L.~Itti, ``State-of-the-art in visual attention modeling,''
  \emph{{IEEE} Trans. Pattern Anal. Machine Intell.}, vol.~35, no.~1, pp.
  185--207, 2013.

\bibitem{micheloni_TCSVT2011}
B.~Dieber, C.~Micheloni, and B.~Rinner, ``Resource-aware coverage and task
  assignment in visual sensor networks,'' \emph{{IEEE} Trans. Circuits Syst.
  Video Technol.}, vol.~21, no.~10, pp. 1424--1437, Oct 2011.

\bibitem{sommerlade2010probabilistic}
E.~Sommerlade and I.~Reid, ``Probabilistic surveillance with multiple active
  cameras,'' in \emph{Proc. {IEEE} ICRA}.\hskip 1em plus 0.5em minus
  0.4em\relax IEEE, 2010, pp. 440--445.

\bibitem{Roy-Chow_TIP2012}
C.~Ding, B.~Song, A.~Morye, J.~Farrell, and A.~Roy-Chowdhury, ``Collaborative
  sensing in a distributed ptz camera network,'' \emph{{IEEE} Trans. Image
  Processing}, vol.~21, no.~7, pp. 3282--3295, July 2012.

\bibitem{Roy-Chow_CST2014}
A.~Morye, C.~Ding, A.~Roy-Chowdhury, and J.~Farrell, ``Distributed constrained
  optimization for bayesian opportunistic visual sensing,'' \emph{{IEEE} Trans.
  Contr. Syst. Technol.}, vol.~22, no.~6, pp. 2302--2318, Nov 2014.

\bibitem{kankanhalli2006expB}
M.~S. Kankanhalli, J.~Wang, and R.~Jain, ``Experiential sampling on multiple
  data streams,'' \emph{{IEEE} Trans. Multimedia}, vol.~8, no.~5, pp. 947--955,
  2006.

\bibitem{martinel_micheloni_2014saliency}
N.~Martinel, C.~Micheloni, and G.~L. Foresti, ``Saliency weighted features for
  person re-identification,'' in \emph{Proc. ECCV}, no.~i, 2014, pp. 1--17.

\bibitem{ejaz2013efficient}
N.~Ejaz, I.~Mehmood, and S.~W. Baik, ``Efficient visual attention based
  framework for extracting key frames from videos,'' \emph{Signal Processing:
  Image Communication}, vol.~28, no.~1, pp. 34--44, 2013.

\bibitem{lee2012discovering}
Y.~J. Lee, J.~Ghosh, and K.~Grauman, ``Discovering important people and objects
  for egocentric video summarization,'' in \emph{Proc. {IEEE} CVPR}, June 2012,
  pp. 1346--1353.

\bibitem{zhao2014quasi}
B.~Zhao and E.~P. Xing, ``Quasi real-time summarization for consumer videos,''
  in \emph{Proc. {IEEE} CVPR}.\hskip 1em plus 0.5em minus 0.4em\relax IEEE,
  2014, pp. 2513--2520.

\bibitem{napoletano2014attentive}
P.~Napoletano and F.~Tisato, ``An attentive multi-camera system,'' in
  \emph{IS\&T/SPIE Electronic Imaging}.\hskip 1em plus 0.5em minus 0.4em\relax
  International Society for Optics and Photonics, 2014, pp. 90\,240O--90\,240O.

\bibitem{bocc08tcsvt}
G.~Boccignone, A.~Marcelli, P.~Napoletano, G.~Di~Fiore, G.~Iacovoni, and
  S.~Morsa, ``{Bayesian integration of face and low-level cues for foveated
  video coding},'' \emph{{IEEE} Trans. Circuits Syst. Video Technol.}, vol.~18,
  no.~12, pp. 1727--1740, 2008.

\bibitem{waage1979foraging}
J.~K. Waage, ``Foraging for patchily-distributed hosts by the parasitoid,
  nemeritis canescens,'' \emph{The J. of Animal Ecology}, pp. 353--371, 1979.

\bibitem{BocFerSMCB2013}
G.~Boccignone and M.~Ferraro, ``Ecological sampling of gaze shifts,''
  \emph{IEEE Trans. Cybernetics}, vol.~44, no.~2, pp. 266--279, 2014.

\bibitem{TatlerBallard2011eye}
B.~Tatler, M.~Hayhoe, M.~Land, and D.~Ballard, ``Eye guidance in natural
  vision: Reinterpreting salience,'' \emph{Journal of vision}, vol.~11, no.~5,
  2011.

\bibitem{BocCOGN2014}
A.~Clavelli, D.~Karatzas, J.~Llad\'{o}s, M.~Ferraro, and G.~Boccignone,
  ``Modelling task-dependent eye guidance to objects in pictures,''
  \emph{Cognitive Computation}, vol.~6, no.~3, pp. 558--584, 2014.

\bibitem{seo2009}
H.~Seo and P.~Milanfar, ``Static and space-time visual saliency detection by
  self-resemblance,'' \emph{J. Vis.}, vol.~9, no.~12, pp. 1--27, 2009.

\bibitem{Poggio2010}
S.~Chikkerur, T.~Serre, C.~Tan, and T.~Poggio, ``What and where: A bayesian
  inference theory of attention,'' \emph{Vis. Res.}, vol.~50, no.~22, pp.
  2233--2247, 2010.

\bibitem{rensink2000dynamic}
R.~Rensink, ``The dynamic representation of scenes,'' \emph{Visual Cognition},
  vol.~1, no.~3, pp. 17--42, 2000.

\bibitem{koller2009probabilistic}
D.~Koller and N.~Friedman, \emph{Probabilistic graphical models: principles and
  techniques}.\hskip 1em plus 0.5em minus 0.4em\relax Cambridge, MA: MIT press,
  2009.

\bibitem{BocBoosted}
G.~Boccignone, P.~Campadelli, A.~Ferrari, and G.~Lipori, ``Boosted tracking in
  video,'' \emph{{IEEE} Signal Processing Lett.}, vol.~17, no.~2, pp. 129--132,
  2010.

\bibitem{TorrJOSA}
A.~Torralba, ``Modeling global scene factors in attention,'' \emph{JOSA A},
  vol.~20, no.~7, pp. 1407--1418, 2003.

\bibitem{mcnamara2006bayes}
J.~M. McNamara, R.~F. Green, and O.~Olsson, ``Bayes' theorem and its
  applications in animal behaviour,'' \emph{Oikos}, vol. 112, no.~2, pp.
  243--251, 2006.

\bibitem{killeen1996bayesian}
P.~R. Killeen, G.-M. Palombo, L.~R. Gottlob, and J.~Beam, ``Bayesian analysis
  of foraging by pigeons (\emph{Columba livia}).'' \emph{J. Exp. Psychology:
  Animal Behavior Processes}, vol.~22, no.~4, p. 480, 1996.

\bibitem{walther2006}
D.~Walther and C.~Koch, ``Modeling attention to salient proto-objects,''
  \emph{Neural Networks}, vol.~19, no.~9, pp. 1395--1407, 2006.

\bibitem{shiner1999}
J.~Shiner, M.~Davison, and P.~Landsberg, ``{Simple measure for complexity},''
  \emph{Physical review E}, vol.~59, no.~2, pp. 1459--1464, 1999.

\bibitem{vanAlphen2008}
J.~J. van Alphen and C.~Bernstein, \emph{Information Acquisition, Information
  Processing, and Patch Time Allocation in Insect Parasitoids}.\hskip 1em plus
  0.5em minus 0.4em\relax Blackwell Publishing Ltd, 2008, pp. 172--192.

\bibitem{denina2011videoweb}
G.~Denina, B.~Bhanu, H.~T. Nguyen, C.~Ding, A.~Kamal, C.~Ravishankar,
  A.~Roy-Chowdhury, A.~Ivers, and B.~Varda, ``Videoweb dataset for multi-camera
  activities and non-verbal communication,'' in \emph{Distributed Video Sensor
  Networks}.\hskip 1em plus 0.5em minus 0.4em\relax Springer, 2011, pp.
  335--347.

\bibitem{itti1998model}
L.~Itti, C.~Koch, and E.~Niebur, ``A model of saliency-based visual attention
  for rapid scene analysis,'' \emph{{IEEE} Trans. Pattern Anal. Machine
  Intell.}, vol.~20, no.~11, pp. 1254--1259, 1998.

\bibitem{lundberg1990functional}
P.~Lundberg and M.~{\AA}str{\"o}m, ``Functional response of optimally foraging
  herbivores,'' \emph{J. Theoret. Biology}, vol. 144, no.~3, pp. 367--377,
  1990.

\bibitem{he2013imbalanced}
H.~He and Y.~Ma, \emph{Imbalanced Learning: Foundations, Algorithms, and
  Applications}.\hskip 1em plus 0.5em minus 0.4em\relax John Wiley \& Sons,
  2013.

\bibitem{dorr2010variability}
M.~Dorr, T.~Martinetz, K.~Gegenfurtner, and E.~Barth, ``Variability of eye
  movements when viewing dynamic natural scenes,'' \emph{J. Vis.}, vol.~10,
  no.~10, 2010.

\bibitem{CucchiaraMobile2010}
R.~Cucchiara and G.~Gualdi, ``\BIBforeignlanguage{English}{Mobile video
  surveillance systems: An architectural overview},'' in
  \emph{\BIBforeignlanguage{English}{Mobile Multimedia Processing}}, ser.
  Lecture Notes in Computer Science, X.~Jiang, M.~Ma, and C.~Chen, Eds.\hskip
  1em plus 0.5em minus 0.4em\relax Springer Berlin Heidelberg, 2010, vol. 5960,
  pp. 89--109.

\bibitem{hall1997introduction}
D.~L. Hall and J.~Llinas, ``An introduction to multisensor data fusion,''
  \emph{Proc. {IEEE}}, vol.~85, no.~1, pp. 6--23, 1997.

\bibitem{driver1998crossmodal}
J.~Driver and C.~Spence, ``Crossmodal attention,'' \emph{Current opinion in
  neurobiology}, vol.~8, no.~2, pp. 245--253, 1998.

\bibitem{tisato2012grounding}
F.~Tisato, C.~Simone, D.~Bernini, M.~P. Locatelli, and D.~Micucci, ``Grounding
  ecologies on multiple spaces,'' \emph{Pervasive and mobile computing},
  vol.~8, no.~4, pp. 575--596, 2012.

\end{thebibliography}

\end{document}